\documentclass[sigconf]{acmart}

\usepackage{xspace}
\usepackage{enumitem}

\usepackage{microtype}
\usepackage{booktabs}
\usepackage{makecell}

\usepackage{array}
\usepackage{tabularx}
\usepackage{multirow}
\usepackage{enumitem}
\usepackage{pifont}
\usepackage{listings}

\lstdefinestyle{prompt}{
    basicstyle=\ttfamily\small,
    backgroundcolor=\color{gray!10},
    frame=single,
    breaklines=true,
    columns=fullflexible,
    keepspaces=true,
    showstringspaces=false
}

\AtBeginDocument{%
  }

\copyrightyear{2026}
\acmYear{2026}
\setcopyright{cc}
\setcctype{by}
\acmConference[KDD '26]{Proceedings of the 32nd ACM SIGKDD Conference on Knowledge Discovery and Data Mining V.2}{August 09--13, 2026}{Jeju Island, Republic of Korea}
\acmBooktitle{Proceedings of the 32nd ACM SIGKDD Conference on Knowledge Discovery and Data Mining V.2 (KDD '26), August 09--13, 2026, Jeju Island, Republic of Korea}
\acmDOI{10.1145/3770855.3817492}
\acmISBN{979-8-4007-2259-2/2026/08}

\begin{document}

\newcommand{\benchname}{AISE-Bench\xspace}
\newcommand{\benchnum}{1,133\xspace}
\newcommand{\testnum}{150\xspace}
\newcommand{\doublechecknum}{250\xspace}
\newcommand{\baselinenum}{14\xspace}
\newcommand{\doubleverifynum}{250\xspace}

\newcommand{\vpara}[1]{\vspace{0.07in}\noindent\textbf{#1 }}

\newcommand{\hide}[1]{}
\newcommand{\todo}[1]{\textbf{\color{red}[(TODO: #1 )]}}  
\newcommand{\zfj}[1]{\textbf{\color{orange}[(ZFJ: #1 )]}}  

\setlength{\abovecaptionskip}{5pt plus 1pt minus 1pt}
\setlength{\textfloatsep}{5pt plus 1.0pt minus 2pt}
\setlength{\dbltextfloatsep}{6pt plus 1pt minus 1pt}

\title{\benchname: A Full-Cycle Curated Benchmark for Information Seeking on Academic Knowledge Graphs}

\author{Fanjin Zhang}
\authornote{Equal contribution.}
\affiliation{%
  \department{School of Information \\Engineering Research Center of Database and Business Intelligence}
  \institution{Renmin University of China }
  \city{Beijing}
  \country{China}}
\email{fanjinz@ruc.edu.cn}

\author{Zhengyang Wang}
\authornotemark[1]
\affiliation{%
  \department{School of Computer Science and Technology}
  \institution{Anhui University}
  \city{Hefei}
  \country{China}}
\email{e125211019@stu.ahu.edu.cn}

\author{Ruixuan Huang}
\affiliation{%
  \department{Z-Lab}
  \institution{Z.ai}
  \city{Beijing}
  \country{China}}
\email{hrx202211@163.com}

\author{Kefan Zhang}
\affiliation{%
  \department{Department of Statistics and Data Science}
  \institution{Tsinghua University}
  \city{Beijing}
  \country{China}}
\email{zkf25@mails.tsinghua.edu.cn}

\author{Amy Xin}
\affiliation{%
  \department{Department of Computer Science and Technology}
  \institution{Tsinghua University}
  \city{Beijing}
  \country{China}}
\email{xin-x25@mails.tsinghua.edu.cn}

\author{Yuanchun Wang}
\affiliation{%
  \department{School of Information \\Key Laboratory of Data Engineering and Knowledge Engineering}
  \institution{Renmin University of China}
  \city{Beijing}
  \country{China}}
\email{wangyuanchun@ruc.edu.cn}

\author{Shu Zhao}
\authornote{Shu Zhao and Juanzi Li are the corresponding authors.}
\affiliation{%
  \department{School of Computer Science and Technology}
  \institution{Anhui University}
  \city{Hefei}
  \country{China}}
\email{zhaoshuzs2002@hotmail.com}

\author{Evgeny Kharlamov}
\affiliation{%
\department{Bosch Center for AI}
  \institution{Renningen, Germany;}
  \institution{\& Department of Informatics}
  \city{University of Oslo}
  \country{Oslo, Norway}}
\email{Evgeny.Kharlamov@de.bosch.com}

\author{Jie Tang}
\affiliation{%
  \department{Department of Computer Science and Technology}
  \institution{Tsinghua University}
  \city{Beijing}
  \country{China}}
\email{jietang@tsinghua.edu.cn}

\author{Juanzi Li}
\authornotemark[2]
\affiliation{%
  \department{Department of Computer Science and Technology}
  \institution{Tsinghua University}
  \city{Beijing}
  \country{China}}
\email{lijuanzi@tsinghua.edu.cn}

\renewcommand{\shortauthors}{Fanjin Zhang et al.}

\begin{abstract}


Large language models (LLMs) augmented with tools are emerging as autonomous agents capable of using Web engine, APIs, and code to solve complex, long‑horizon tasks. Current tool-using benchmarks for information seeking on academic graphs rely on synthetic templates, simplified solution spaces, or narrow tasks such as paper-centric tasks, leaving key challenges underexplored -- realistic user intent, complex multi‑step API planning, rich parameter filling for APIs, grounded answers with references, and comprehensive evaluation of both the process and the outcome. We introduce AISE-Bench, a real-world, full-cycle annotated benchmark for information seeking on academic knowledge graphs. AISE-Bench release contains 1,133 QA pairs, including query taxonomies, full API execution trajectories, validated parameters, and source-grounded answers with reference links. To support high‑quality annotation, we design a customized agent workflow to enable annotators to plan, execute, and revise complex API workflows efficiently. We develop a comprehensive evaluation protocol measuring answer quality, reference grounding, API‑planning correctness, and execution success. Among the 14 evaluated methods, even the strongest model (PLAY2PROMPT with Gemini‑3‑Pro) achieves only moderate performance and often struggles with API planning and execution. AISE-Bench establishes a challenging new testbed for quantitatively evaluating and improving the stepwise correctness, grounded summarization, and traceable reasoning of multi-step API-using LLM agents. Our code and data are available\footnote{\url{https://aise-bench.github.io/}}.



  
\end{abstract}

\ccsdesc[500]{Information systems~Information retrieval}

\hide{

\ccsdesc[500]{Do Not Use This Code~Generate the Correct Terms for Your Paper}
\ccsdesc[300]{Do Not Use This Code~Generate the Correct Terms for Your Paper}
\ccsdesc{Do Not Use This Code~Generate the Correct Terms for Your Paper}
\ccsdesc[100]{Do Not Use This Code~Generate the Correct Terms for Your Paper}
}

\keywords{Academic Search, LLM Agent, Academic Knowledge Graph}


\maketitle

\section{Introduction}

Foundation models are now able to leverage tool‑calling mechanisms (e.g., search~\cite{jin2025search}, coding~\cite{jimenez2024swe}, and APIs~\cite{qin2024toolllm}) to substantially elevate their upper bounds of performance in complex and  long‑horizon tasks.
Notably, these models have showcased the potential as intelligent agents in general domain such as GUI agents~\cite{xu2025mobilerl}, Web agents~\cite{he2024webvoyager}, and multi‑agent social simulation~\cite{wang2025user}. 

\begin{table*}[t]
    \newcolumntype{C}{>{\centering\arraybackslash}c}
    \centering
    \newcolumntype{P}[1]{>{\centering\arraybackslash}p{#1}}
    \begin{tabular}{c|@{~ }CP{0.6cm}CCCC}
            \toprule[1.2pt]
        {Benchmarks}  &  {Real Queries}  & {Entity} & {Synthetic Data} & {Taxonomy} & {Evaluation Metrics} & {Annotation Modules}  \\ 
        \midrule
        PeerQA~\cite{baumgartner2025peerqa} & \textcolor{green}{{\ding{51}}}   & papers & \textcolor{red}{\ding{55}} & - & correctness & answer text, evidence  \\
        ScholarQABench\cite{asai2024openscholar} & \textcolor{green}{{\ding{51}}} & papers & \textcolor{red}{\ding{55}} & tasks, disciplines & LLM, citations & answer text, citations \\ 
        SoAyBench~\cite{Wang_2025} & \textcolor{red}{\ding{55}}  & multi. & \textcolor{green}{{\ding{51}}} & solution library & process and answer & -   \\
        DeepDive~\cite{lu2025deepdiveadvancingdeepsearch} & \textcolor{red}{\ding{55}}  & multi. & \textcolor{green}{{\ding{51}}} & - & - & -  \\
        \midrule
        \benchname & \textcolor{green}{{\ding{51}}} & multi. & \textcolor{red}{\ding{55}} & intent, disciplines & LLM, process, citations & API paths, answer, citations  \\
        \bottomrule[1.2pt]
    \end{tabular}
    \vspace{0.2cm}
    \caption{Comparison of academic search Benchmarks. \textnormal{
    The \textit{Entity} column denotes the type of academic entity targeted by each QA task (multi. = multiple).
    In the \textit{Evaluation Metrics} column, LLM is LLM-based semantic correctness, citations indicates that the references or links used in the answer need to be evaluated, and process indicates reasoning process evaluation.
    In the \textit{Annotation Modules} column, evidence indicates that the supporting context for the answer needs to be located in the original text. Citations indicates that the answer must provide its referenced sources. API paths indicates that the API call traces, including the API inputs and outputs, need to be annotated.}}
    \label{table:comparison_bench}
\end{table*}

Although LLM tool use has advanced rapidly, existing benchmarks fall short of addressing \textit{real and specialized} academic questions that require  \textit{complex input parameters and API planning}. As shown in Table \ref{table:comparison_bench}, SoAyBench~\cite{Wang_2025} constructs <query, solution, code> triplets via template generation and augmentation.
DeepDive~\cite{lu2025deepdiveadvancingdeepsearch} synthesizes complex reasoning paths via random walks over academic graphs.  PeerQA~\cite{baumgartner2025peerqa} and ScholarQABench~\cite{asai2024openscholar} focus on the paper understanding, largely ignoring other entities such as authors and venues.
However, at the \textit{query level}, prior arts rarely reflect real user interactions with academic knowledge graphs, 
leading to biased query distributions and misalignment with human information needs. 
In \textit{API planning}, many methods rely on preset simplified solution spaces, 
limiting exploratory reasoning.
At the \textit{answer level}, free‑form answers in existing datasets make it difficult to discern whether there are supporting sources.

Thus, we present \benchname, a \textbf{full-cycle annotated, real-world} API-using benchmarks  for specialized information seeking over academic knowledge graphs. It provides APIs for entity search, entity detail querying, and entity relationship querying across papers, authors, venues, and organizations. 
The benchmark is built from complex, real academic questions collected from AMiner~\cite{tang2008arnetminer}, and enables comprehensive evaluation of LLMs in intent understanding, API planning, complex parameterization, and source-grounded, faithful summarization over long contexts. 

\benchname includes \benchnum rigorously annotated, multi-disciplinary questions spanning diverse entity types and knowledge levels, including \doublechecknum double-reviewed and 883 single-reviewed instances.
We design a custom agent workflow 
that streamlines annotation, enabling one-click execution of full API workflows or fine-grained editing of individual API calls.
Beyond queries and answers, \benchname includes query taxonomies, API trajectories, and grounded reference links, supporting comprehensive evaluation of reasoning processes, answer quality, and citation accuracy.

Accordingly, we propose a comprehensive evaluation of 
API-using LLMs using metrics that assess answer quality, reference accuracy, API planning, and execution success. 
Experiments span \baselinenum state-of-the-art methods, including 6 LLMs, 4 API-using agent frameworks, 2 coding agents, and 2 commercial deep research systems. 
The results show that even the best model, PLAY2PROMPT with Gemini-3-Pro, achieves only $61.04\%$ correctness and $60.9\%$ completeness as measured by LLM judges, 
highlighting persistent limitations in understanding, planning, and executing specialized academic queries. Our key contributions are as follows: 


\begin{itemize}[leftmargin=*]
    \item We introduce \benchname, a real-world, full-cycle-annotated, API-using benchmarks for information seeking over academic graphs, featuring queries with taxonomic types, API trajectories, answers with embedded reference links.
    \item We propose an elaborate agent framework to assist annotation via seamlessly suggesting API planning, executing single or multiple API calls, and summarizing answers.
    \item We develop an extensive evaluation suite measuring reference‑link matching performance, answer accuracy, API planning, parameter accuracy, and execution success rate.
    \item We conduct an in-depth evaluation of \baselinenum methods, revealing clear limitations in complex API planning, API execution, and disambiguation of similar APIs.
\end{itemize}

 \section{Related Work}

\subsection{LLM API-Using Methods}

LLM API-using methods can be categorized into \textbf{large foundation models, training-free agent frameworks, and training-based methods.}
Many flagship LLMs now natively support multi‑turn tool invocation, including GPT~\cite{singh2025openai}, DeepSeek~\cite{liu2025deepseek}, Gemini~\cite{comanici2025gemini}, etc.


Training‑free tool‑use methods enable LLMs to invoke external tools without additional model training.
A representative paradigm is ReAct \cite{yao2022react}, which interleaves natural‑language reasoning with action execution, allowing models to iteratively plan, call tools, and update beliefs based on observations. 
Another line relies on prompt engineering to elicit reliable zero‑shot or few‑shot tool invocation, as exemplified by methods such as DRAFT \cite{qu2025exploration}, PLAY2PROMPT \cite{fang2025play2prompt}, SoAy \cite{Wang_2025}, AvaTaR \cite{wu2024avataroptimizingllmagents}, TURA~\cite{zhao2025tura}, and CodeAct \cite{wang2024executablecodeactionselicit}. 
For instance, PLAY2PROMPT~\cite{fang2025play2prompt} enables true zero‑shot tool use by automatically probing tools to refine documentation and generate examples without labeled data. 

In contrast, training‑based approaches improve tool use via model optimization. 
Supervised fine‑tuning methods such as ToolLLM~\cite{qin2024toolllm}, Granite~\cite{abdelaziz2024granite}, and ToolGen~\cite{wang2025toolgen} enhance tool adaptation using curated tool‑calling traces.
Reinforcement learning-based methods further optimize tool selection and action planning, including ToolRL~\cite{qian2026toolrl}, ReTool~\cite{feng2025retool}, Tool‑Star~\cite{dong2025tool}, and FunRL~\cite{hao2025exploring}, achieving notable gains in accuracy and decision quality~\cite{Qu_2025}.

\begin{figure*}[t]
\centering
\includegraphics[width=1.0\linewidth]{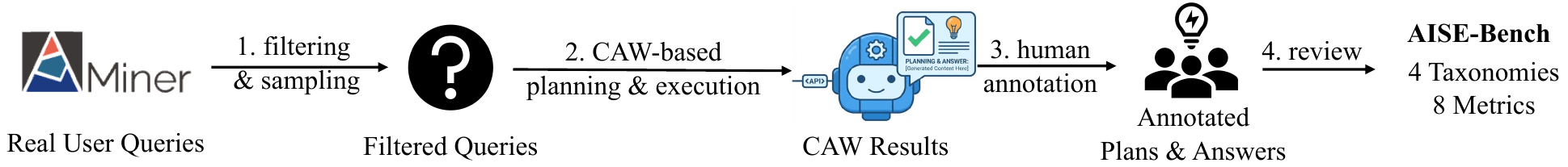} 
\caption{\benchname construction pipeline. We first perform filtering and sampling based on real user queries from AMiner. Then, using a customized agent workflow (CAW), we generate initial API plans and answers for each query. Annotators refine the CAW-based plans and answers, and each annotated query is verified by at least one reviewer.}
\label{fig:bench_construction}
\end{figure*}

\subsection{Academic Search Benchmarks}

To enhance large foundation models’ ability to seek academic information, understand scholarly literature, and use complex academic APIs, some academic search benchmarks~\cite{li2024dalkdynamiccoaugmentationllms,chen2026rpc,baumgartner2025peerqa,Wang_2025,patel2025deepscholar,singh2025ai2,zhang2024oag} have recently emerged.

These benchmarks can be grouped by the capabilities they target. 
PeerQA~\cite{baumgartner2025peerqa}, RPC‑Bench~\cite{chen2026rpc}, Scholar QA~\cite{singh2025ai2}, and the PaperQA~\cite{lala2023paperqa} focus on paper‑centric evaluation, emphasizing retrieval, grounding, and scientifically rigorous question answering (QA) over academic corpora. 
The DeepResearch family of benchmarks~\cite{du2025deepresearch,wan2026deep}, together with DeepScholar‑Bench~\cite{patel2025deepscholar}, provides more holistic evaluations of research workflow, including problem decomposition, and specialized tasks such as generating related‑work sections. In contrast, LitSearch~\cite{ajith2024litsearch} targets complex literature search queries.
More relevant to our work, SoAy~\cite{Wang_2025} and DeepDive~\cite{lu2025deepdiveadvancingdeepsearch} 
construct QA tasks over academic knowledge graphs using predefined paths or random walks.



However, existing related benchmarks have notable limitations: 
(1) they primarily focus on paper understanding and retrieval, with limited support for API composition and invocation over academic knowledge graphs;
(2) academic API‑using benchmarks often rely on automatically constructed trajectories and oversimplified API libraries and solution spaces; and
(3) their evaluations are narrow, typically assessing only short‑answer accuracy. 

To address these limitations, we introduce \benchname. At the \textit{method level}, it targets complex API composition over academic knowledge graphs. At the \textit{problem and planning level}, we select 
real user queries and a rich API library, with detailed annotations including question types, multi-step API‑calling trajectories, and reference‑grounded answers. 
In terms of \textit{evaluation}, \benchname 
supports comprehensive evaluation of reference usage, API planning and execution, and answer correctness and completeness.



\section{\benchname Construction}
\label{sec:bench_construction}

The objective of \benchname is to assess LLMs’ ability to perform multi-hop information seeking over academic knowledge graphs in realistic user scenarios.
Starting from real academic search queries, we build a comprehensive API library over an academic knowledge graph and design a full-cycle annotation framework covering query classification, intent understanding, API planning and execution, and citation-grounded answer generation. \benchname enables fine-grained evaluation of multi-step reasoning and planning, accurate API parameterization, and trustworthy, reference-backed summarization.
Figure \ref{fig:bench_construction} shows the overall construction framework. 


\hide{
AIS-Bench (\textbf{AI Academic Intelligent Search Benchmark}) evaluates tool-augmented academic search and question answering in realistic settings.
In contrast to benchmarks focusing solely on retrieval quality or single-turn QA, AIS-Bench targets the full interaction loop:
\emph{intent understanding $\rightarrow$ tool planning $\rightarrow$ API execution $\rightarrow$ evidence-grounded synthesis $\rightarrow$ citation formatting}.
This enables diagnosis of failures in multi-step planning, API parameterization, grounding, and verifiable citation generation.
}

\subsection{Data Collection and Processing Pipeline}
\label{subsec:ais-data}

We firstly collect real user queries from academic search system AMiner~\cite{tang2008arnetminer}. 
After removing extremely short, overly long, and corrupted queries, we conduct preliminary annotation along four dimensions:
(1) API solvability: whether the query can be answered through a combination of API calls in the API Library;
(2) Task complexity, indicating whether multi‑step reasoning ($\geq 4$  API calls) is required; 
(3) Entity‑related user intent, including searching for papers, authors, organizations, and venues;
(4) Knowledge level:
(4.1) Knowledge memorization: no need for summarization or transformation; the answer can be directly presented based on the retrieved entities.
(4.2) Knowledge understanding: requires interpretation and synthesis retrieved information, including:
\begin{itemize}[leftmargin=*]
    \item Comparison: involving multiple entities (papers, authors, etc.)
    \item Examples: providing concrete instances or applications
    \item Interpretation: explaining or analyzing the underlying knowledge
    \item Summarization: offering structured or high‑level summaries
\end{itemize}

From the preliminary annotations, 
we select questions addressable by our APIs and apply stratified sampling to ensure coverage across problem types. 
The following sections describe the API library and annotation guidelines.

\subsection{API Library}

\begin{figure*}[h]
\centering
\includegraphics[width=1.0\linewidth]{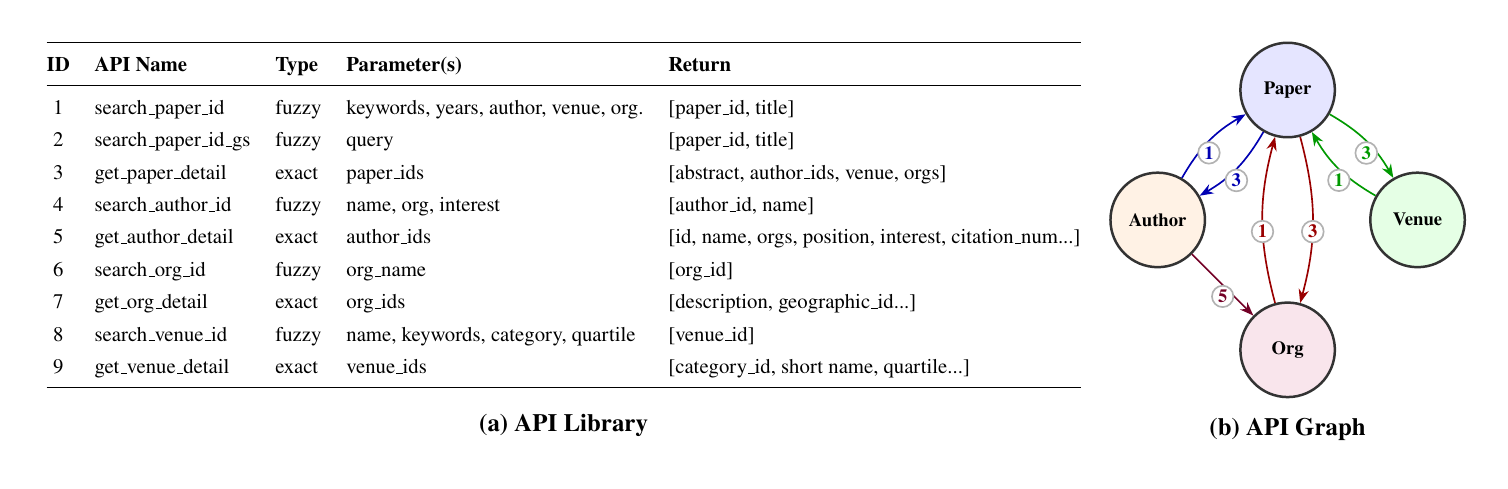} 
\caption{\textbf{Overview of the API ecosystem.} \textnormal{(a) \textbf{API Library}: Detailed specifications of available API functions, including their types, input parameters, and return values. search\_paper\_id\_gs is Google Scholar Search. (b) \textbf{API Graph}: Schematic representation of the interactions between core entities (Paper, Author, Venue, and Org), where numbered labels correspond to the API IDs defined in the library.}}
\label{fig:api_library}
\end{figure*}

We consider four core entity types in academic knowledge graphs: papers, authors, venues, and organizations.
Our API library provides dedicated endpoints for each entity type, grouped into entity search APIs and entity detail APIs.
Search APIs return candidate entity IDs based on textual queries, while detail APIs provide rich attributes and inter-entity relations (e.g., author-paper and paper-venue links) for a given ID. 
In total, we include nine APIs. For paper retrieval, we integrate both the AMiner and Google Scholar search APIs to improve coverage. Figure \ref{fig:api_library} 
summarizes the API definitions and supported entity relationships. 


Although our API library contains nine endpoints over four academic entity types, AISE-Bench is designed to be representative and extensible rather than exhaustive. The current API space is intentionally compact so that gold trajectories remain executable, comparable, and well-defined. It captures key structural properties of real academic KGs, including heterogeneous search and detail APIs, multi-hop inter-entity relations, and partially overlapping data sources. Google Scholar provides broad paper-search coverage, while AMiner provides richer scholarly entity APIs for authors, venues, and organizations. Since both platforms aggregate multi-source scholarly records and perform entity disambiguation~\cite{zhang2019oag}, the nine API categories cover the main interaction patterns needed for academic information seeking. The overlap between AMiner and Google Scholar also creates nontrivial API-selection and source-disambiguation challenges.

In addition, adding many highly overlapping APIs may not necessarily improve the benchmark, because multiple equally valid API choices would make annotation and evaluation more ambiguous and make it harder to distinguish genuine planning errors from alternative valid trajectories. Our framework is API-agnostic and can be readily extended to other academic KGs (e.g., OpenAlex~\cite{priem2022openalex}) or other domains. Therefore, we view the API library as a representative testbed rather than an exhaustive enumeration.
\subsection{Customized Agent Workflow}
\label{subsec:custom_agent_workflow}

\begin{figure}[t]
\centering
\includegraphics[width=1.0\linewidth]{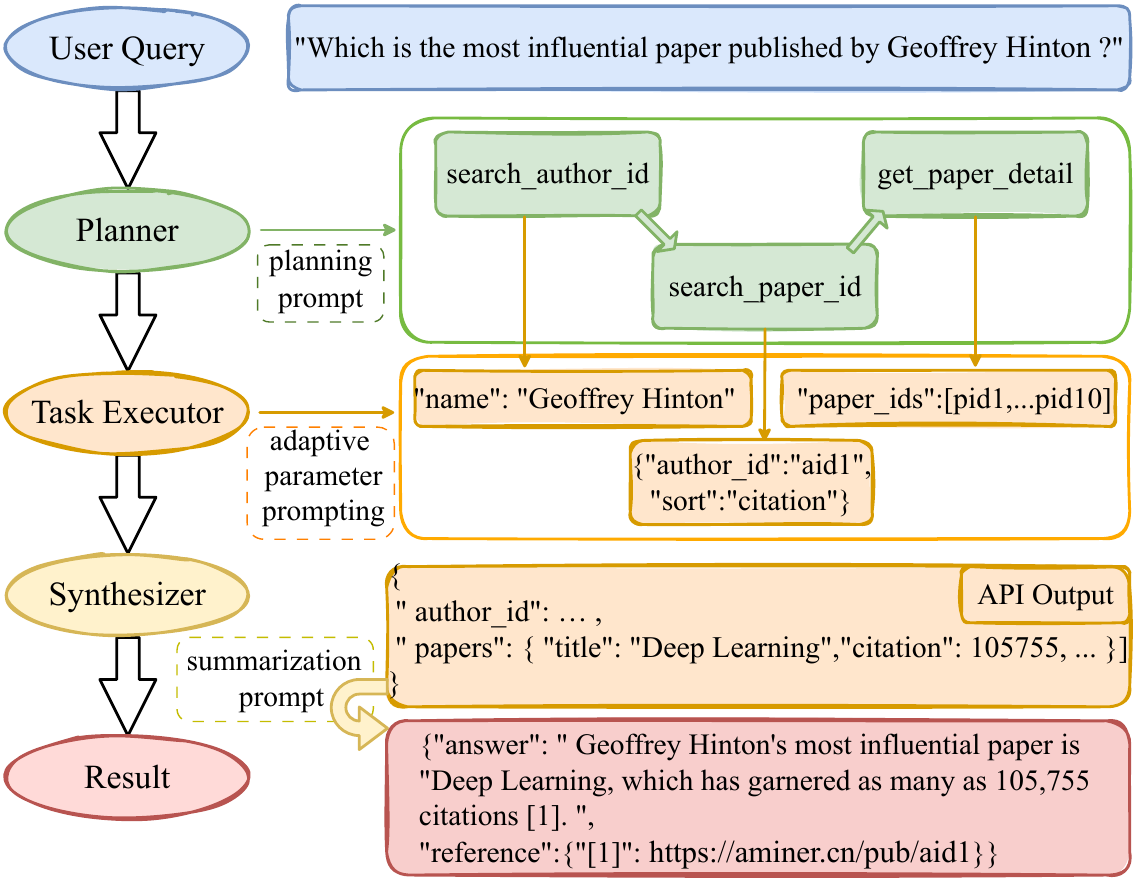} 
\caption{Overview of the Customized Agent Workflow (CAW) framework, which consists of three main components: a planner, a task executor, and a synthesizer. The modular multi-agent design allows annotators to independently modify or customize individual modules.}
\label{fig:caw_framework}
\end{figure}

\begin{figure*}[h]
\centering
\includegraphics[width=1.0\linewidth]{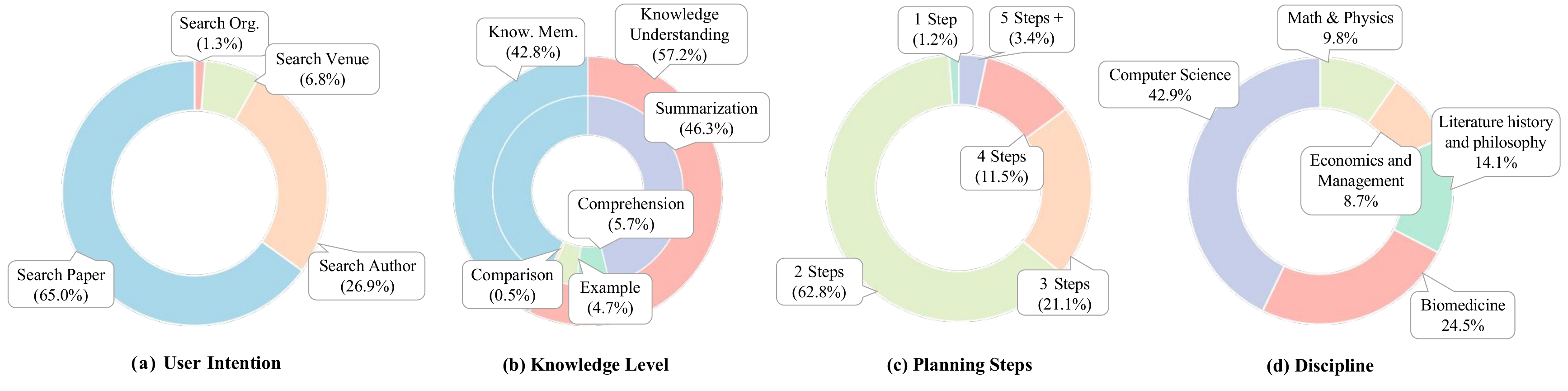} 
\caption{Distribution of annotated questions across four dimensions: 
    \textnormal{
    (a) \textbf{User Intention}: Search Org. = Search Organization; 
    (b) \textbf{Knowledge Level}: Know. Mem. = Knowledge Memorization. Knowledge Understanding is further categorized into Examples, Comparison, Comprehension, and Summarization; 
    (c) \textbf{Planning Steps}: number of steps of API calls.
    (d) \textbf{Discipline}: first-level discipline.
    }}
\label{fig:questions_category}
\end{figure*}

We propose a three‑stage Customized Agent Workflow (CAW) that decomposes complex queries into executable retrieval operations over an academic knowledge graph. As depicted in Figure \ref{fig:caw_framework}, CAW comprises  a \textbf{Planner}, \textbf{Task Executor}, and \textbf{Synthesizer}. Given a user query, the Planner uses an LLM to generate a step‑wise API execution plan structured as a directed acyclic graph (DAG). 
Each DAG node represents an API call with specified inputs, dependencies, and execution order. 
This design enables flexible multi‑hop retrieval (e.g., resolving an author entity before retrieving publications) without manual rule engineering.

The Task Executor schedules tasks asynchronously under dependency constraints, executing independent tasks in parallel and deferring dependent ones until prerequisites complete. 
When required parameters are unavailable at execution time (e.g., entity IDs from prior outputs), the executor invokes an LLM to dynamically instantiate them using API‑specific argument templates, decoupling plan generation from execution and enabling flexible adaptation to diverse query intents.

We design two prompt templates for plan generation and answer synthesis. 
The planning prompt casts the LLM as an API planning expert that selects from nine predefined academic APIs, each with fixed input-output schemas, enforcing strict JSON-only outputs and default constraints (e.g., English keywords and citation-count sorting). 
The synthesis prompt treats the LLM as a domain expert that generates concise, evidence-grounded answers strictly based on retrieved results. 
All factual claims are supported by inline citations linked to entity-specific URLs\footnote{e.g., https://aminer.cn/pub/\{paper\_id\} }. 
The final output is a structured JSON containing the synthesized answer and references.



\subsection{Annotation Process}

We develop an interactive, API‑based annotation system with a structured evaluation workflow (see Appendix \ref{sec:anno_system}). The pipeline includes user login, API debugging, model‑generated planning, stepwise execution, and final summarization. When the CAW framework successfully produces and executes a valid API plan, the user only reviews and optionally edits the generated plan and output. Otherwise, the user intervenes by manually defining API trajectories, specifying parameters, and executing the steps.


Annotators can flexibly adjust API parameters and test calls via a debugging interface that displays input parameters and responses. The annotation workflow centers on a planning‑and‑execution module, where annotators can generate multi‑step plans using a reasoning model, manually edit plans, or execute them in one click. Individual API calls may also be run independently with user‑specified parameters. All executions are logged, including API inputs and outputs. 



The summary stage enforces strict validation: outputs must follow a predefined JSON schema with answer and reference fields, include correctly paired inline citations, use non-fabricated links, and maintain relevance and internal consistency between inline citations and references.


The goal of this annotation protocol is to manage the full lifecycle of answering academic questions, including API planning, parameter specification, execution, and citation‑based answer synthesis. This enables transparent and reproducible evaluation of both process and results. 



\subsection{Quality Control}

We perform rigorous annotation review with strict quality control. Each answer is first checked for compliance with the required JSON format, including both the response and supporting links. Content accuracy, completeness, and verifiability via citations are then assessed. API plans are thoroughly reviewed by expanding all inputs and outputs to ensure parameter completeness and validity. Every instance is reviewed by at least one domain expert, and 
\doubleverifynum 
instances undergo double verification.


When an annotation was deemed incorrect, reviewers provided explicit justifications to enable a second round of correction. We evaluated 356 samples involving complex answers and reasoning processes, and achieved an inter‑reviewer agreement of $0.6608$, indicating a reliable and consistent review process.



In total, we annotated over 6,000 candidate instances and completed more than 4,000 reviews, resulting in \benchnum question-answer pairs that passed the review process. We observe that the pass rate remains relatively low because professional academic API-using questions require full-cycle annotation, including taxonomy labeling, executable API trajectories, parameter validation, grounded answer writing, citation verification, and expert review. We categorize questions by discipline, entity-centric intent, knowledge level, and API-planning complexity. The distribution across these dimensions is shown in Figure ~\ref{fig:questions_category}. The \testnum test questions used in the main experiments are selected from the double-reviewed subset via stratified sampling to cover disciplines, entity-centric intents, knowledge levels, and API-planning complexity. We release the questions, question categories, answers with links, and API trajectories. AMiner API is accessible via token-based authentication for research purposes.

\subsection{Evaluation Metrics}
\label{subsec:eval_metrics_desc}

We propose a multi‑dimensional evaluation framework covering reasoning, answer quality, and references. For \textbf{API calling and planning}, we measure planning graph edit distance, parameter accuracy, and execution success rate. For \textbf{references and formatting}, we compare model‑generated citations with gold annotations using precision and recall, and assess compliance with the required JSON output format. For \textbf{answer content}, we evaluate correctness and completeness, analogous to precision and recall against reference answers, and measure faithfulness by detecting inconsistencies with the API outputs.


\vpara{API-based Judge.} \textit{Planning Graph Edit Distance.}
We measure the discrepancy between predicted and gold API-call plans using graph edit distance. Each plan is converted into a DAG of API names based on execution order, and the distance is computed between the predicted and reference graphs, capturing differences in both API selection and ordering. For some questions, multiple API trajectories may lead to the same correct answer. We therefore interpret graph edit distance as a proxy for structural alignment with a concise reference plan, rather than as a perfect measure of planning optimality. Its value is to capture API-selection errors, ordering errors, and unnecessary planning overhead. 


\textit{Parameter Accuracy.} 
We evaluate predicted API parameters against gold inputs by aligning the union of API steps in the prediction and reference, assigning zero to missing or extra steps. For matched steps, scalar parameters require exact matches, while list parameters are scored by element-level overlap. Step scores are averaged to obtain question-level accuracy, and the final metric is the mean accuracy over all questions. We additionally report a fuzzy parameter-level F1 score in Appendix~\ref{app:additional_process_metrics} to account for normalized semantic matches between predicted and gold API arguments.

\textit{Execution Success Rate.} 
It measures the reliability of multi-step tool execution. A task scores 1 if all steps complete without error, and 0 otherwise. Higher values indicate more robust end-to-end API-execution ability. We further report Partial Completion Score in Appendix~\ref{app:additional_process_metrics} to measure partial execution progress in long-horizon API trajectories.


\vpara{References and Formatting.} \textit{Precision and Recall:}
We extract all URLs from the gold and predicted answer strings and measure URL accuracy and coverage via precision and recall. Although exact URL matching can be strict for arbitrary webpages, AISE-Bench normalizes references to canonical entity URLs for papers, authors, venues, and organizations. Since AMiner performs multi-source data fusion and entity disambiguation~\cite{zhang2019oag}, different versions of the same paper are largely merged into a single entity ID. Thus, this metric is closer to canonical entity-link matching than arbitrary webpage matching.

\textit{Format metric}: The output must be a single JSON object with exactly two fields, answer and reference. Reference field must be a nonempty dictionary of consecutively numbered citations ([1], [2], …) mapped to valid URLs, and the answer must use exactly the same citation set, with no extras or omissions.


\vpara{Answer Content.} \textit{Correctness} 
is an LLM‑based semantic metric that evaluates factual alignment between a model’s free‑form output and a gold reference (akin to precision). Unlike URL‑based metrics, it captures nuanced meaning by prompting an LLM judge to assign a score in [0,1] reflecting the proportion of accurate and relevant information in the prediction. High scores mean the model’s answers are mostly correct, while low scores show the answers contain content that cannot be verified by the annotated answers.

\textit{Completeness} 
is an LLM‑based content‑coverage metric that measures how fully a model’s output captures the key information in a gold reference. An LLM evaluator assigns a score in [0,1] reflecting the fraction of essential reference content present in the prediction, with higher scores indicating more complete coverage and lower scores indicating missing information. 

\textit{Faithfulness} 
measures the extent to which a model’s answer is strictly grounded in the provided API output, assessed via an LLM‑as‑a‑judge that scores the presence and severity of unsupported or hallucinated content on a continuous [0,1] scale.

\textit{F1-LM} is the harmonic mean of correctness and completeness, balancing accuracy and coverage.



\hide{
We construct AIS-Bench from real user questions and an academic search backend (AMiner), following a multi-stage pipeline:
(i) collect $40{,}000+$ real queries from research scenarios;
(ii) clean queries by removing overly short/long or garbled inputs;
(iii) label each query by \textbf{intent} and \textbf{discipline} (top-level fields);
(iv) perform stratified sampling over \emph{intent $\times$ discipline} to build a traceable evaluation set of roughly $2{,}000$ questions with an approximate Train:Test split of 3:1;
(v) bootstrap candidate answers and tool plans with LLMs, then apply human editing and selection to produce gold annotations.

So far, we have manually verified $500+$ instances containing \texttt{question}, \texttt{result\_edit} (final structured answer),
\texttt{planning\_text} (API plan), and the associated \texttt{api\_input}/\texttt{api\_output} traces.
Each instance is validated by at least one domain-proficient reviewer, and 180 instances are double-verified.
}

\hide{
\subsection{Taxonomy}
\label{subsec:ais-taxonomy}

AIS-Bench uses a workflow-aligned taxonomy.

\paragraph{Discipline distribution (10,000+ analyzed queries).}
\begin{itemize}
  \item Computer Science: 6142
  \item Computing and Information Science: 1488
  \item Engineering: 1060
  \item Education: 326
  \item Business: 132
  \item Social Science: 57
  \item Environment: 56
\end{itemize}

\paragraph{Intent distribution (10,000+ analyzed queries).}
\begin{itemize}
  \item Find papers: 6675
  \item Find scholars: 2969
  \item Find venues (journals/conferences): 475
  \item Find institutions: 218
  \item Find patents: 49
\end{itemize}

\paragraph{Question depth.}
We distinguish \textbf{knowledge recall} (directly listing retrieved entities) and \textbf{knowledge understanding} (requiring synthesis), including comparison, example generation, interpretation, and summarization.

\paragraph{Planning difficulty.}
We define planning difficulty by the number of API calls in the execution trace; queries requiring $\geq 4$ calls are labeled as complex.

}

\hide{
\subsection{Task Definition and Output Format}
\label{subsec:ais-task}

Given an input question, the system produces a structured output stored as \texttt{result\_edit}.
AIS-Bench requires the final answer to be \textbf{machine-checkable} and \textbf{citation-grounded}:

\begin{verbatim}
{
  "answer": "Natural language answer with inline citations [1], [2], ...",
  "reference": {
    "[1]": "https://www.aminer.cn/...",
    "[2]": "https://www.aminer.cn/..."
  }
}
\end{verbatim}

\noindent The benchmark additionally stores:
(i) \texttt{planning\_text}: an ordered list of tool/API calls (with step order and tool name),
(ii) \texttt{api\_input}: serialized inputs to each call,
(iii) \texttt{api\_output}: serialized outputs returned by the tools.
These artifacts enable evaluation of both \emph{planning correctness} and \emph{answer grounding}.
}

\hide{
\subsection{Evaluation Protocol (Aligned with Our Scoring Implementation)}
\label{subsec:ais-eval}

We implement a hybrid evaluation protocol combining deterministic programmatic checks and LLM-based judging.
Let $G$ denote the gold instance and $P$ the prediction for the same question id (\texttt{qid}).
Gold and prediction each contain: \texttt{result\_edit}, \texttt{planning\_text}, and (for predictions) \texttt{api\_input}/\texttt{api\_output}.

\subsubsection{Automatic Metrics}
\label{subsubsec:auto}

\paragraph{Reference Precision and Recall.}
We extract all URLs from the gold and predicted \texttt{result\_edit} strings using a regular expression.
Let $L_G$ and $L_P$ be the sets of extracted URLs.
We define:
\begin{align}
\mathrm{Precision} &= 
\begin{cases}
\frac{|L_P \cap L_G|}{|L_P|}, & |L_G|>0 \wedge |L_P|>0 \\
0, & \text{otherwise}
\end{cases}
\\
\mathrm{Recall} &= 
\begin{cases}
\frac{|L_P \cap L_G|}{|L_G|}, & |L_G|>0 \wedge |L_P|>0 \\
0, & \text{otherwise.}
\end{cases}
\end{align}

\paragraph{Clarity (Schema Validity).}
We score \texttt{Clarity} $\in \{0,1\}$ by verifying that the predicted \texttt{result\_edit} contains \emph{only} a JSON object (optionally wrapped in Markdown fences).
The JSON must have exactly two fields: \texttt{answer} and \texttt{reference}.
The \texttt{reference} must be a non-empty dictionary whose keys match the pattern \texttt{[n]} with continuous numbering starting from \texttt{[1]},
and each value must include the AMiner domain \texttt{https://www.aminer.cn}.
Finally, \texttt{answer} must contain \emph{exactly} the same set of citation markers as the keys of \texttt{reference} (no missing and no extra markers).

\paragraph{API Execution Success.}
We set \texttt{Success} $\in \{0,1\}$ to 1 if both \texttt{api\_input} and \texttt{api\_output} are JSON-serializable (i.e., can be round-tripped via \texttt{json.dumps} and \texttt{json.loads}); otherwise 0.

\paragraph{Planning Edit Distance.}
We compute a Levenshtein edit distance between the predicted and gold API call sequences.
Each plan is converted into an ordered list of tool names (sorted by the \texttt{order} field), and we report:
\begin{equation}
\mathrm{EditDistance} = \mathrm{Lev}\big(\langle \mathrm{name}(P_i)\rangle,\ \langle \mathrm{name}(G_j)\rangle\big).
\end{equation}
This measures mismatches in tool selection and ordering.

\subsubsection{LLM-as-a-Judge Metrics}
\label{subsubsec:judge}

We use an external judge model to output a scalar score $\in [0,1]$ in strict JSON format \texttt{\{"rating": x\}}.
To improve stability, the judge is queried at temperature 0.3 and the rating is clipped to $[0,1]$.

\paragraph{Correctness.}
Given question $q$, gold answer $a_G$ and predicted answer $a_P$, the judge scores whether $a_P$ accurately answers $q$ and matches the substance of $a_G$.
If $a_P$ only describes how to search (procedures) without providing results, the score is forced to 0.

\paragraph{Integrality (Coverage vs.\ Gold).}
The judge scores the completeness of $a_P$ relative to $a_G$, penalizing missing key information points.
If $a_P$ only describes search steps without results, the score is forced to 0.

\paragraph{Task Completion.}
To assess whether the response fulfills the user's \emph{core goal}, the judge:
(i) decomposes $q$ into sub-queries (if applicable),
(ii) identifies the central intent for each sub-query,
(iii) extracts an evaluation checklist from $a_G$ (ignoring reference URLs),
and then scores how well $a_P$ satisfies the checklist.

\paragraph{Faithfulness (Grounding to API Output).}
The judge scores whether $a_P$ is strictly supported by the predicted \texttt{api\_output} (penalizing fabricated or hallucinated claims).

}

\hide{
\subsection{Implementation Notes}
\label{subsec:impl}

For missing predictions (no matching \texttt{qid}), we assign zeros to all metrics.

\paragraph{Important caveat.}
Our current Precision/Recall implementation extracts URLs from the raw \texttt{result\_edit} text (not the parsed \texttt{reference} dictionary).
This makes link matching robust to formatting but may underestimate scores if a system uses citation markers without explicit URLs in the text.
In future versions, we plan to compute link-level Precision/Recall directly from the parsed \texttt{reference} field to better align with the benchmark format.

\begin{table}[t]
\centering
\small
\begin{tabular}{p{2.6cm}p{4.9cm}p{1.4cm}}
\hline
\textbf{Metric} & \textbf{What it measures} & \textbf{Range} \\
\hline
Precision & URL overlap fraction: $|L_P \cap L_G|/|L_P|$ & $[0,1]$ \\
Recall & URL overlap fraction: $|L_P \cap L_G|/|L_G|$ & $[0,1]$ \\
Clarity & Strict JSON schema + AMiner URL + citation-key match & $\{0,1\}$ \\
Success & JSON-serializability of \texttt{api\_input}/\texttt{api\_output} & $\{0,1\}$ \\
EditDistance & Levenshtein distance over ordered tool-name sequences & $\mathbb{N}$ \\
Correctness & Judge score for answer accuracy vs.\ question and gold & $[0,1]$ \\
Integrality & Judge score for information coverage vs.\ gold & $[0,1]$ \\
Completion & Judge score for fulfilling core goal via checklist & $[0,1]$ \\
Faithfulness & Judge score for grounding to API outputs & $[0,1]$ \\
\hline
\end{tabular}
\caption{AIS-Bench evaluation dimensions aligned with our scoring implementation.}
\label{tab:ais-metrics}
\end{table}
}
\section{Experiments}

\begin{table*}[t]
\small
\centering
        \renewcommand{\arraystretch}{1.2}  
        \begin{tabular}{c|c|c c c|c c c|c c c c}  
            \toprule[1.1pt]
        \multirow{2}{*}{\shortstack{Model\\Type}}    
            &\multirow{2}{*}{Model } 
            &\multicolumn{3}{c|}{References and Formatting}
            &\multicolumn{3}{c|}{API-based Judge}
            &\multicolumn{4}{c}{Answer Content}
        \\
        \cmidrule{3-5} \cmidrule{6-8}\cmidrule{9-12}
        & & {Precision} & {Recall} & {Format} & {Edit Dist.}  &{Para. Acc.}&{Success}  & {Correct.} & {Complete.} & {Faithful.} & {F1-LM} \\
        \midrule
        \multirow{6}{*}{CAW}
        & Deepseek-V3.2   & 0.3544 & 0.3461 & 0.78 & 1.56 & \textbf{0.4453} & 0.8267 & 0.4571 & 0.4729 & 0.8355 & 0.4649  \\
        & GLM-4.7         & 0.1905 & 0.1659 & 0.4067 & 1.8467 & 0.3474 & 0.8533 & 0.3727 & 0.3510 & 0.6168 & 0.3615 \\
        & Qwen3-235B-A22B           & \textbf{0.4416} & 0.3524 & \textbf{0.8467} & 1.84 & 0.4131 & 0.9133 & 0.4936 & 0.4778 & 0.7607 & 0.4856 \\
        & GPT-5.2         & 0.3008 & 0.3167 & \textbf{0.8467} & 5.4667 & 0.3432 & 0.62 & 0.4368 & 0.4562  & 0.787 & 0.4463 \\
        & Gemini-3-Pro    & 0.4109 & 0.4342 & 0.74 & \textbf{1.2867} & \underline{0.4242} & 0.7867 & 0.5721 & 0.5495 & 0.7907 & 0.5606 \\
        & Claude-4.5      & 0.1564 & 0.1072 & 0.1467 & 1.7733 & 0.3632 & 0.7733 & 0.3666 & 0.3328 & 0.6705 & 0.3489 \\

        \midrule
        \multirow{4}{*}{\shortstack{API-Using\\Agent}}
        & ReAct        & 0.343 & 0.3779 & 0.7333  & 4.8267 & 0.2705 & \textbf{0.9933} & 0.5923 & \underline{0.6015} & 0.7402 & \underline{0.5969} \\
        & AvaTaR       & \underline{0.4313} & \underline{0.4639} & 0.79  & 1.3867 & 0.3522 & 0.9267 & \underline{0.6046} & 0.5894 & 0.826 & \underline{0.5969} \\
        & DRAFT        & 0.4199 & 0.4545 & 0.7667 & 1.3333 & 0.412 & 0.92 & 0.5873 & 0.5819 & 0.8217 & 0.5846 \\
        & PLAY2PROMPT  & 0.4308 & \textbf{0.4881} & \underline{0.8333} & 1.5267 & 0.3968 & 0.9 & \textbf{0.6104} & \textbf{0.609} & \underline{0.8542} & \textbf{0.610} \\
        \midrule
        \multirow{2}{*}{\shortstack{Coding\\Agent}}
        & CodeAct & 0.4022 & 0.4313 & 0.8 & 1.3467 & 0.4047 & 0.9467 & 0.5144 & 0.5123 & \textbf{0.9295}  & 0.5130 \\
        & SoAy    & 0.4275 & 0.4306  & 0.8067  & \underline{1.3067} & 0.3934 & \underline{0.9667} & 0.541 & 0.5008 & 0.7225  & 0.5201 \\
        \midrule
        \multirow{2}{*}{\shortstack{Deep Research\\Agent}}
        & Perplexity & / & / & / & / & / & / & 0.3692 & 0.4251 & / & 0.3952 \\
        & Metaso     & / & / & / & / & / & / & 0.2688 & 0.3025 & / & 0.2847 \\
        \bottomrule[1.2pt]
    \end{tabular}
    \vspace{0.2cm}
    \caption{Main evaluation results on the test set. \textnormal{Edit Dist. = Planning Graph Edit Distance. Para. Acc. = Parameter Accuracy. Success = Execution Success Rate. Correct. = Correctness. Complete. = Completeness. Faithful. = Faithfulness. }}
    \label{tb:main_results}
\end{table*}

\subsection{Experimental Setup} 
\label{subsec:exp_setup}

We construct a test set of \testnum questions, each independently verified by two reviewers. The baselines span four categories:


(1) \textbf{CAW}: 
employs a customized multi-agent workflow (Sec.~\ref{subsec:custom_agent_workflow}) with specialized agents for API planning, execution, and summarization. We evaluate CAW on several frontier LLMs: DeepSeek-V3.2~\cite{liu2025deepseek}, GLM-4.7~\cite{glm2024chatglm}, Qwen3-235B-A22B~\cite{yang2025qwen3}, GPT-5.2~\cite{singh2025openai}, Claude-Sonnet-4.5\footnote{\url{https://www.anthropic.com/claude/sonnet}}, and Gemini-3-Pro~\cite{comanici2025gemini}.

(2) \textbf{API-using agents}: 
\textit{ReAct}~\cite{yao2022react} interleaves reasoning and actions to enable LLMs to plan and interact with external environments. \textit{AvaTaR}~\cite{wu2024avataroptimizingllmagents} optimizes tool usage by contrastive prompt refinement, while \textit{DRAFT}~\cite{qu2025exploration} improves tool documentation through iterative LLM feedback and trial‑and‑error. \textit{PLAY2PROMPT}~\cite{fang2025play2prompt} enables zero‑shot tool usage by probing tools to refine documentation and generate examples without labeled data.

(3) \textbf{Coding agents}: 
CodeAct \cite{wang2024executablecodeactionselicit} provides a unified action framework that allows LLM agents to generate and execute Python code, overcoming the rigidity of fixed action schemas and limited toolsets.
SoAy \cite{Wang_2025} is a solution‑oriented approach that guides LLMs to invoke academic APIs via code generation following predefined API‑coupling sequences, simplifying API interactions and improving reasoning efficiency.

(4) \textbf{Deep Research agents}: 
\textit{Perplexity Deep Research}\footnote{\url{https://www.perplexity.ai}} performs multi‑step, iterative research by combining web search, code execution, and reasoning to produce expert‑level answers.
\textit{Metaso}\footnote{\url{https://metaso.cn/search-api/playground}} is an MCP‑based search and QA system that supports real‑time web retrieval and RAG‑powered responses.

\vpara{Evaluation Metrics.} As described in Section \ref{subsec:eval_metrics_desc}, we evaluate answers using multiple metrics. These metrics capture different perspectives of output quality and are organized into three groups. Except for the planning graph edit distance, other metrics are normalized to [0,1]. For deep research systems without exposed APIs or execution traces (e.g., Perplexity, Metaso), we evaluate only answer-level metrics (Correctness and Completeness).

\hide{
\begin{itemize}[leftmargin=*]
    \item Citation‑based metrics include \textit{Precision}, \textit{Recall}, and \textit{Format}, which measure how well the generated content aligns with the ground‑truth citations and whether the output conforms to the required format.
    \item API‑usage metrics include \textit{Edit Distance} and \textit{Execution Success Rate}, which assess how closely the generated API calls align with the expected API planning graph and whether the planned API sequence is successfully executed to completion.
    \item LLM‑as‑a‑judge metrics include \textit{Correctness}, \textit{Completeness}, and \textit{Faithfulness}, which capture higher‑level semantic quality as assessed by LLMs. We also compute the average of these three metrics (denoted as \textit{Avg.}) to provide an overall measure of answer quality.
\end{itemize}
}

\vpara{Implementation Details.}
For both API‑using and coding agents, we use Gemini‑3‑Pro, the strongest LLM available in the CAW framework, as the base model. For LLM‑as‑a‑judge, we consider the top three CAW models and select the two whose judgments best align with human preferences; the final score is the average of their judge scores. We incorporate an automatic retry mechanism to enhance fault tolerance in the CAW framework. We adopt an exponential backoff strategy to mitigate resource contention from concurrent retries, thereby enhancing system stability.




\subsection{Main Results} 
Table \ref{tb:main_results} summarizes the performance of LLM agent frameworks on the \benchname test set. Traditional reference‑based metrics substantially underestimate answer quality, yielding consistently lower scores than LLM‑as‑a‑judge evaluations of answer content. 

LLMs exhibit substantial performance variability. GLM‑4.7, Qwen3, and DeepSeek-V3.2 achieves strong API execution reliability but underperforms Gemini‑3‑Pro in LLM‑as‑a‑Judge answer quality. GPT‑5.2 performs poorly overall, with the highest Planning Graph Edit Distance and lowest Execution Success Rate, because they tend to plan overly long API paths. In contrast, Gemini‑3‑Pro leads on Correctness, Completeness, and Faithfulness, demonstrating superior accuracy and robustness. 


API‑using agent frameworks improve performance across evaluation dimensions. PLAY2PROMPT achieves the highest F1-LM and CodeAct attains the best faithfulness. ReAct, AvaTaR, and DRAFT also produce good-quality answers. Overall, improved tool‑interaction logic and response structuring enhance alignment between LLMs and academic APIs. In contrast, deep research agents underperform in correctness, completeness, and faithfulness, underscoring the importance of richer tool libraries. 


Current models struggle to perform well across all evaluation dimensions, with none achieving consistently strong results. Additional fine-grained process metrics further reveal the source of execution failures. As shown in Appendix~\ref{app:additional_process_metrics}, most methods achieve high partial completion scores (0.82-0.95), but much lower fuzzy parameter F1 scores (0.15-0.42). This gap indicates that many models can complete most API steps, while accurate parameter filling remains a major bottleneck. AvaTaR, CodeAct, and DRAFT lead on partial completion, whereas DeepSeek-V3.2 achieves the best fuzzy parameter F1. These findings highlight persistent challenges in jointly optimizing semantic accuracy, information coverage, and API planning and execution.

\subsection{Results across Question Types}

\begin{figure}[h]
\centering
\includegraphics[width=0.9\linewidth]{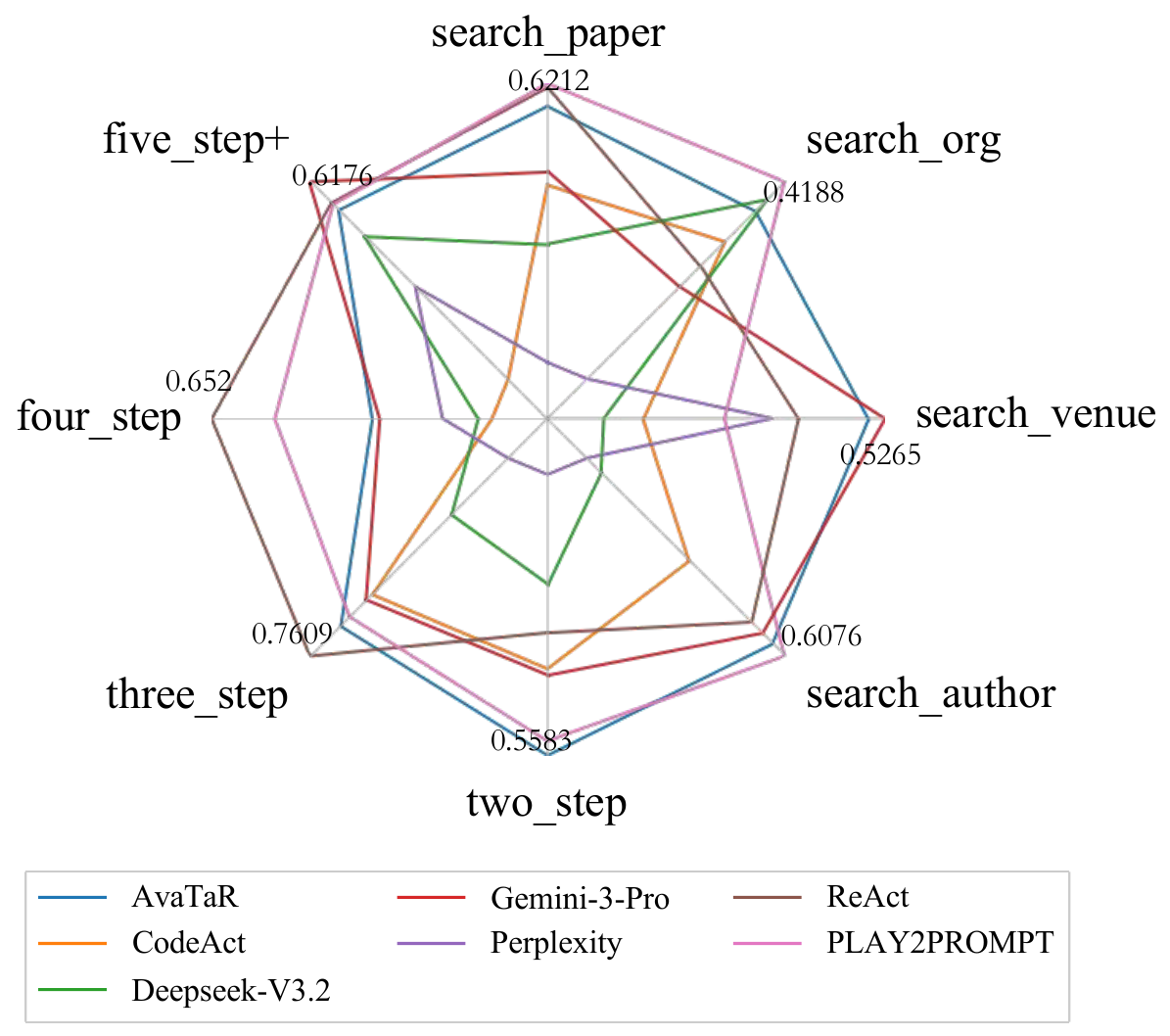} 
\caption{The F1-LM score of representative methods on different types of problems (user intent understanding and API planning difficulty).}
\label{fig:question-type-analysis}
\end{figure}

We analyze how several representative methods perform on different types of questions from the perspectives of user intent, API‑planning difficulty, and knowledge level.

\vpara{Entity-centric User Intent.}
Figure \ref{fig:question-type-analysis} shows performance variation across entity-centric intents, including search papers (\textit{search\_paper}), search authors (\textit{search\_author}), search venues (\textit{search\_venue}), and search organizations (\textit{search\_org}).

AvaTaR and PLAY2PROMPT perform strongly on common tasks such as paper and author search. For less frequent tasks, including organization and venue search, Gemini‑3‑Pro, PLAY2PROMPT, and AvaTaR generalize best. Overall, AvaTaR demonstrates robust planning, execution, and summarization abilities enabled by few‑shot and contrastive reasoning. In contrast, the Perplexity Deep Research agent underperforms across most tasks, achieving competitive results only in venue search, likely due to the higher accessibility of venue information compared to the complexity of paper understanding and author profiling.

\vpara{API Planning Difficulty.}
We bucket queries by the order of the final API call, yielding sequences of 2 to 5+ steps, where \textit{five\_step+} reflects long-horizon reasoning. Gemini-3-Pro performs best in the \textit{five\_step+} setting (0.6176), highlighting its robust long-context modeling and multi-step planning. CodeAct is competitive on short-horizon tasks but exhibits an approximately linear degradation as sequence length increases, indicating error accumulation in long-horizon execution. ReAct shows a non-monotonic trend, peaking at mid-range tasks (three\_step: 0.7609; four\_step: 0.652) before dropping sharply on \textit{five\_step+} tasks (0.5766), suggesting limitations of iterative reasoning-action loops under deep dependency chains. Perplexity Deep Research performs poorly across all settings, with minimal sensitivity to planning depth.

\begin{figure}[h]
\centering
\includegraphics[width=1.0\linewidth]{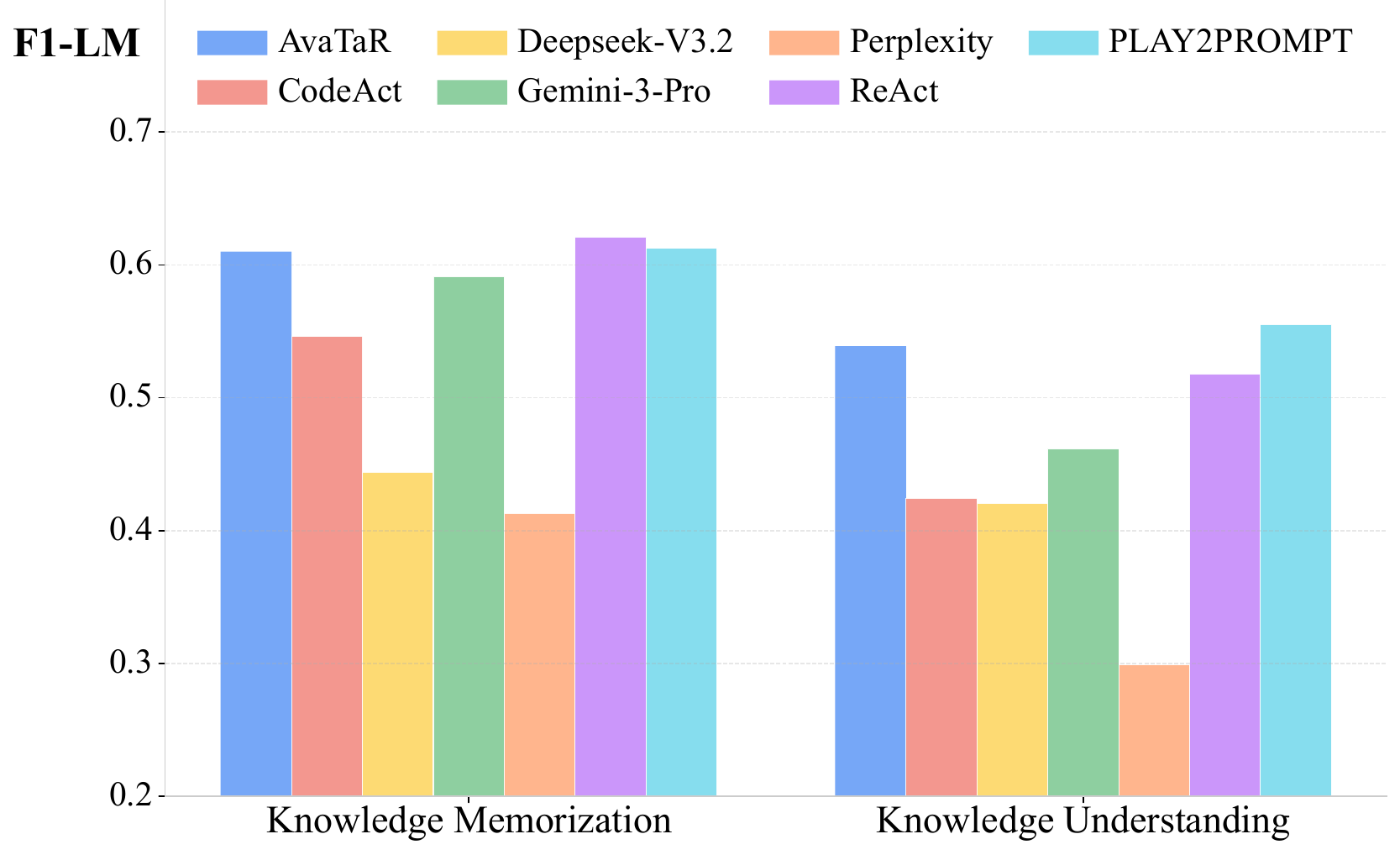} 
\caption{The performance of representative methods on queries of knowledge memorization and understanding.}
\label{fig:results_knowledge_level}
\end{figure}

\vpara{Knowledge Level.}
We analyze model performance across increasing question complexity by distinguishing knowledge memorization from higher-order understanding. As shown in Figure~\ref{fig:results_knowledge_level}, all methods perform consistently better on memorization queries, with a clear performance gap that widens as cognitive demands increase.

For knowledge memorization, ReAct achieves the best performance, followed by PLAY2PROMPT and AvaTaR. Gemini-3-Pro performs comparably, suggesting strong inherent knowledge acquisition across methods. DeepSeek-V3.2 shows weaker results, likely due to information loss from sparse attention in long contexts.


\begin{figure*}[t]
\centering
\includegraphics[width=0.98\linewidth]{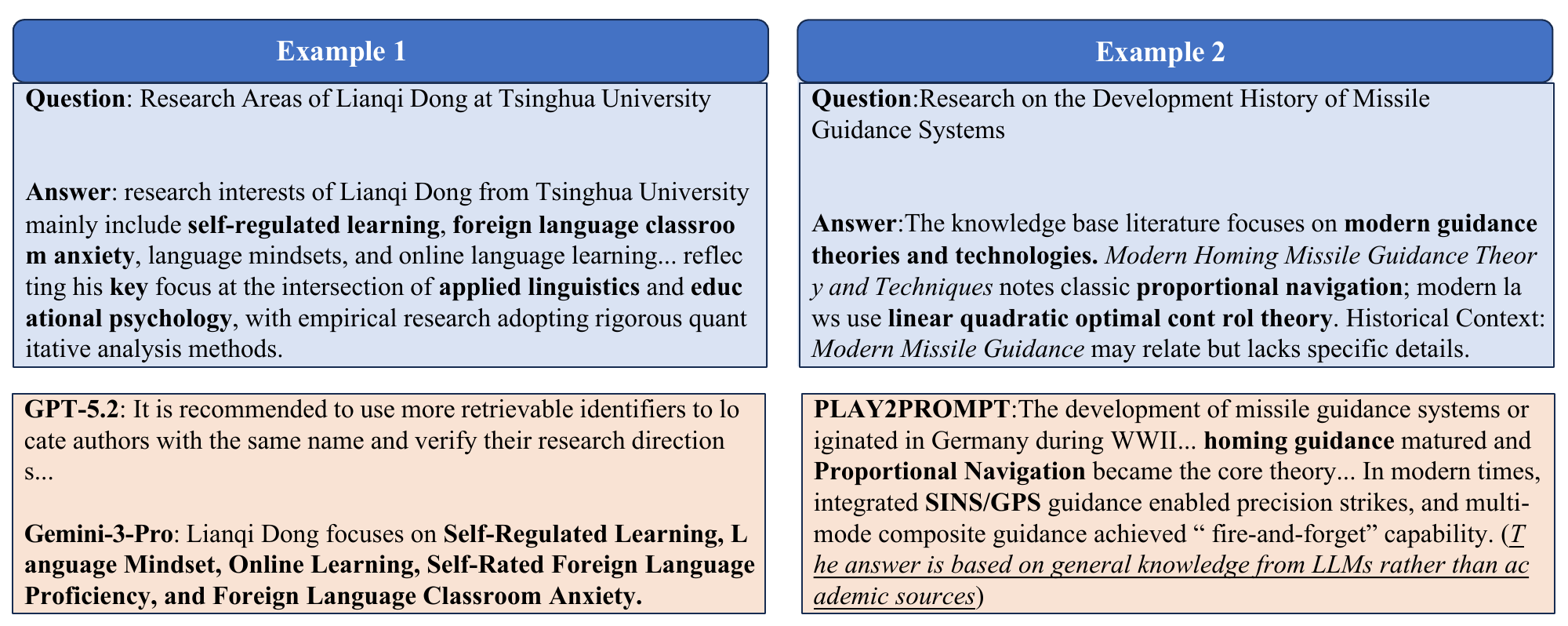} 
\caption{Representative case studies from the \benchname test set. 
}
\label{fig:case_study} 
\end{figure*}

For knowledge-understanding queries, all models degrade due to increased reasoning and summarization demands. CodeAct and ReAct drop by nearly 10\%, while PLAY2PROMPT and AvaTaR are more robust, suggesting superior integration of retrieval and semantic reasoning. Perplexity performs poorly across both knowledge memorization and understanding tasks, highlighting persistent limitations in professional knowledge retrieval and reasoning.

\subsection{Human Judge Consistency}

\begin{table}[htbp]
  \centering
  \begin{tabular}{lccc}
    \toprule
    & P-BT & PW-AUC & Avg. \\
    \midrule
    DeepSeek-V3.2 & \underline{0.7581} & \underline{0.7320} & \underline{0.7451} \\
    Qwen3-235B-A22B & 0.7376 & 0.7239 & 0.7308 \\
    Gemini-3-Pro & \textbf{0.8122} & \textbf{0.7735} & \textbf{0.7929} \\
    \bottomrule
  \end{tabular}
  \vspace{0.1cm}
  \caption{Analysis of agreement between LLM-based judges and human assessments.}
  \label{tab:judge_metrics}
\end{table}

To align LLM judgments with human assessments, we sample open-ended academic QA instances from the test set and generate predictions using the baselines in Section~\ref{subsec:exp_setup}. To control for length bias, we construct 173 pairwise comparisons by selecting outputs with similar lengths. During annotation, output order is randomized and model identities are masked.


We select three strong judge candidates: DeepSeek-V3.2, Qwen3, and Gemini-3-Pro. Each judge is given full context (queries and evaluation criteria) and independently scores alignment with human preferences. Consistency is evaluated using P-BT and PW-AUC (pairwise AUC). P-BT fits a Bradley-Terry model~\cite{turner2012bradley} to pairwise judgments to obtain scalar scores, which are then correlated with human preferences using Pearson correlation.


As shown in Table~\ref{tab:judge_metrics}, Gemini-3-Pro achieves the highest consistency with human judgments, followed by DeepSeek and Qwen. To balance accuracy and robustness, we adopt Gemini-3-Pro and DeepSeek-V3.2 as joint judges for LLM-as-a-Judge evaluations.

\subsection{Case Studies}

We conduct case studies from two perspectives: (1) common failure modes, (2) limitations of different types of methods.

\vpara{Common Failure Modes.} 
Figure \ref{fig:case_study} displays common failure modes (additional cases in Appendix \ref{sec:app_case_studies}), identifying the following representative issues. 
\textbf{(1) Invalid retrieval and broken links} (Example 1): Some models return empty results or invalid references. For example, DeepSeek‑V3.2 retrieves no valid author records for ``scholars engaged in gender studies in mainland China'', while GPT‑5.2 produces broken links when queried about ``the research interest of Lianqi Dong from Tsinghua University'', undermining verifiability. 
\textbf{(2) Insufficient fidelity and overgeneralization} (Example 2): In the absence of valid citations, models may generate generic content. Gemini’s response to ``tax avoidance or tax planning'' and PLAY2PROMPT’s response to ``the development history of missile guidance systems'' lack supporting references. 
(3) \textbf{Disambiguation of Similar APIs.} When facing similar paper-search APIs, the baseline tends to select the more commonly mentioned AMiner API in the prompt. However, for topics requiring in-depth investigation (e.g., Evaluation of Industrial Chain and Supply Chain Resilience), results retrieved via Google Scholar Search are more relevant (Example 3 in Appendix \ref{sec:app_case_studies}).

\vpara{Limitations of Different Types of Methods.}
We evaluate multiple foundation models using customized agent workflow frameworks (Appendix \ref{sec:app_case_studies}) and observe clear performance disparities. Strong models (e.g., Gemini-3-Pro, Qwen3-235B-A22B) exhibit robust topic decomposition and logical reasoning, while weaker models (e.g., GPT-5.2, Claude-Sonnet-4.5) show poor retrieval adaptability and redundant outputs. Notably, GPT-5.2 often fails to return valid retrieval results or usable references for author-centric queries. Across API-using frameworks, we observe substantial variation in structured output and task adaptability. CodeAct consistently produces standardized, academically appropriate formats, whereas DRAFT performs well on topic survey tasks (e.g., chronological analyzes of missile guidance systems). However, most frameworks suffer from low retrieval fidelity and limited evidence utilization, with some responses lacking valid references and relying heavily on general knowledge.

\section{Conclusion}


We present \benchname, a full-cycle annotated  benchmark targeting LLM capabilities in real academic information‑seeking scenarios over large‑scale academic knowledge graphs. The benchmark features authentic queries, fine‑grained question taxonomies, multi-step executable API trajectories, and reference‑grounded answers, enabling systematic assessment of intent understanding, multi‑step tool planning, references, and long‑context synthesis. We further introduce an agent‑assisted annotation workflow named CAW and an evaluation protocol that supports efficient annotation and detailed analysis of execution errors. Experiments on \baselinenum state-of-the-art agent frameworks reveal persistent weaknesses in handling specialized academic queries requiring precise API planning, parameterization, and grounded reasoning. We expect \benchname to support future multi‑step tool‑using LLMs, advancing research toward accurate, comprehensive, and faithful AI agents for scientific knowledge exploration.

\hide{
\begin{acks}
To Robert, for the bagels and explaining CMYK and color spaces.
\end{acks}
}

\begin{acks}

This work has been supported by the National Natural Science Foundation of China (62406164, 62476003, 62476150, 62425601),
Anhui Province Excellent Scientific Research and Innovation Team (2024AH010004), Anhui Provincial Natural Science Foundation - Water Science Joint Fund (2408055US006),
a grant from the Institute for Guo Qiang, Tsinghua University (2019GQB0003),
and New Cornerstone Science Foundation through the XPLORER PRIZE.
We also acknowledge the support from Zhipu AI-Anhui University Joint Research Center, and the High-Performance Computing Platform of Anhui University.
We also acknowledge the support from Public Computing Cloud, Renmin University of China.
It was also partially supported by the
EU Project SMARTY (GA 101140087).

\end{acks}

\bibliographystyle{ACM-Reference-Format}
\bibliography{ref}

\appendix


\section{Ethical Considerations}
All source materials for the benchmark---including academic literature, API documentation, and model interaction data---are derived from publicly available resources (\textit{e.g.}, AMiner, Google Scholar, official tool repositories). No private, confidential, or proprietary information is included, and the use of source content complies with their terms of service and copyright policies. We do not claim ownership of original scholarly texts or tool documentation.

\vpara{Use of LLM Tools}: We use LLM to polish this paper.

\hide{
\subsection{Limitations}
While the AI Academic Search Benchmark offers comprehensive coverage of core academic search scenarios within its scope, several aspects warrant further refinement and expansion:

First, in terms of data annotation, due to the high cost of manual construction of academic QA pairs and tool interaction trajectories, we prioritize manual annotation for the test set to ensure evaluation robustness, while the training portion incorporates LLM-refined annotations (aligned with original academic literature and API return logic) for flexible use by researchers. Future work will increase the proportion of manual annotation to enhance the reliability of training data.

Second, the current benchmark primarily focuses on single-document retrieval, framework comparison, and basic tool call tasks, with limited coverage of complex scholarly scenarios such as cross-domain literature synthesis, multi-round interactive search, and long-text review generation (e.g., ``Related Work'' composition). This narrow task scope reflects the initial focus on core capabilities, and future extensions will include cross-document reasoning and dynamic academic information integration to adapt to real-world research needs.

Third, the test data and model configurations exhibit domain bias: the evaluated tasks and reference literature are predominantly derived from computer science and AI-related fields, with insufficient coverage of life sciences, social sciences, and other disciplines. Additionally, the tool use paradigms tested focus on mainstream frameworks (e.g., DRAFT, ReTool) and lack evaluation of emerging low-resource tool chains or domain-specific tools. The established evaluation pipeline and metric system, however, lay a solid foundation for future cross-disciplinary expansion and tool ecosystem updates.

Fourth, the benchmark's current focus is on static evaluation of model performance, with limited analysis of dynamic adaptability---such as how models adjust tool call strategies for time-sensitive academic information (e.g., newly published papers) or handle API version updates. Future work will introduce dynamic test sets and real-time tool interaction scenarios to address this gap.

\subsection{Ethical Considerations}
This work adheres to the ACL Code of Ethics and follows strict ethical guidelines throughout the benchmark's construction and release:

All source materials for the benchmark---including academic literature, API documentation, and model interaction data---are derived from publicly available resources (\textit{e.g.}, arXiv, OpenReview, official tool repositories). No private, confidential, or proprietary information is included, and the use of source content complies with their terms of service and copyright policies. We do not claim ownership of original scholarly texts or tool documentation.

All released artifacts, including the model test results, evaluation metrics code, and annotation guidelines, are distributed under the Creative Commons Attribution 4.0 International (CC BY 4.0) license, with clear attribution to original authors and source platforms. The benchmark is intended solely for non-commercial academic research purposes, such as model benchmarking, methodological development, and performance analysis, and must not be used for commercial deployment or production-level services without prior authorization.

To protect privacy and prevent misuse, we perform strict data anonymization: removing author names, institutional affiliations, and residual metadata from reference literature and API interaction logs that could enable re-identification. We also conduct rigorous quality control to filter low-quality, irrelevant, or redundant content, and apply balanced sampling to mitigate systematic biases in data sources (\textit{e.g.}, overrepresentation of specific research fields). The benchmark contains no harmful, discriminatory, or security-sensitive content, and complies with global academic ethics and data protection regulations.

No experiments in this work involve personal health data, sensitive demographic attributes, or controversial research topics. Potential conflicts of interest, including institutional affiliations and project sponsorships, are fully disclosed in accordance with academic conference policies. This study is dedicated to advancing the ethical and effective application of AI in academic search, and does not promote or enable malicious use of technology.
}

\section{Additional Process Metrics}
\label{app:additional_process_metrics}
We report two additional process-level metrics to complement the main evaluation. Partial Completion Score (PCS) measures the fraction of API steps that are successfully completed in a multi-step trajectory, providing partial credit for long-horizon plans instead of treating execution as an all-or-nothing outcome. Fuzzy Parameter F1 complements exact parameter accuracy by normalizing case, punctuation, morphology, aliases, list order, and cross-lingual variants such as Chinese and English names of persons and institutions. It evaluates semantic matching over API argument key-value pairs after aligning predicted and gold calls by API name and call index.
Table~\ref{tab:additional_process_metrics} reports additional process-level metrics, including Partial Completion Score (PCS) and Fuzzy Parameter F1. PCS measures partial execution progress, while Fuzzy Parameter F1 evaluates parameter matching after normalization.

\begin{table*}[t]
\centering
\caption{Additional process-level evaluation results. PCS denotes Partial Completion Score, and Fuzzy Param. F1 denotes Fuzzy Parameter F1.}
\label{tab:additional_process_metrics}
\resizebox{\textwidth}{!}{
\begin{tabular}{lcccccccccccc}
\toprule
Metric & AvaTaR & CodeAct & DRAFT & SoAy & PLAY2PROMPT & Claude-4.5 & DeepSeek-V3.2 & Gemini-3-Pro & GLM-4.7 & GPT-5.2 & Qwen3-235B-A22B & ReAct \\
\midrule
PCS & 0.9540 & 0.9515 & 0.9346 & 0.8970 & 0.8985 & 0.8259 & 0.8974 & 0.8543 & 0.8486 & 0.8206 & 0.9210 & 0.9011 \\
Fuzzy Param. F1 & 0.4094 & 0.3919 & 0.3788 & 0.3847 & 0.3790 & 0.3822 & 0.4245 & 0.3897 & 0.3778 & 0.1740 & 0.4062 & 0.1483 \\
\bottomrule
\end{tabular}
}
\end{table*}

Most methods achieve high PCS but much lower Fuzzy Parameter F1, indicating that accurate parameter filling remains a major bottleneck even when models complete most API steps. AvaTaR, CodeAct, and DRAFT lead on partial completion, while DeepSeek-V3.2 achieves the best Fuzzy Parameter F1.

\section{Additional Case Studies}
\label{sec:app_case_studies}
\textbf{Example 3 (Inaccurate API calls):} 
\\\\
\textbf{Question:} Evaluation of Industrial Chain and Supply Chain Resilience
\\\\
\textbf{Referenced Answer:} The core purpose of researching industrial chain and supply chain resilience evaluation is to grasp its research status, evaluation methods and applications. Key findings: Studies generally build evaluation systems based on resilience capabilities..., with various combined quantitative and qualitative methods adopted, including fuzzy logic, integrated decision models, graph theory approach and fuzzy-based hypothesis testing.... Empirical applications have been carried out in industries like automotive, coal and garment..., providing relevant results and suggestions. Future research needs to focus on supply chain complexity differences and integrate emerging technologies..., while Industry 4.0's contribution to KPIs and sustainability is also concerned.... In summary, the research has formed a multi-dimensional, multi-method pattern, deepening in practice and integrating with technology.
\\\\
\textbf{AvaTaR Answer:} From the knowledge base, there are no direct empirical papers on industrial chain and supply chain resilience evaluation, but highly cited literature offers key methodological and theoretical support. Useful methodological frameworks for such research: Systematic review \& index construction—PRISMA guidelines support scientific indicator screening.... Empirical model verification—SEM is a core, with Anderson et al.’s "two-step method" as a classic paradigm.... Data bias control—Podsakoff et al.’s research elaborates on remedies for Common Method Biases in questionnaire data.... Theoretical perspective—Peter Senge’s "learning organization" and "system thinking" lay a theoretical foundation for supply chain resilience....
\\\\
In this case, AvaTaR failed to retrieve direct empirical papers on industrial chain and supply chain resilience evaluation, primarily because AMiner has limitations in resource coverage or relevance matching for this specific subtopic—its indexed papers may focus more on general fields and struggle to accurately target empirical research related to "resilience evaluation". At this point, the search paper id gs (Google Scholar API) should be prioritized. Leveraging its more comprehensive global academic resource coverage and more refined topic correlation algorithms, it can supplement the retrieval of empirical papers directly focusing on this topic, instead of relying solely on methodological frameworks, thereby enhancing the relevance and information completeness of the answer.

\vspace{0.2cm}

\vpara{Example 4 (No academic links available.):} 
\\\\
\textbf{Question:} I want to study "John King Fairbank's Research on China's Political System"
\\\\
\textbf{Referenced Answer:} The user intends to understand John King Fairbank’s research achievements and academic contributions on China's political system. His core research areas included History and the Cultural Revolution, with studies covering modern and contemporary Chinese history, laying the foundation for analyzing China's political system evolution.... His representative works include The Cambridge History of China, China: A New History (systematically interpreting political system changes) and The Great Chinese Revolution, 1800-1985.... His widely cited works laid an important foundation for Western understanding of China's history and political system.... Notably, there is no monograph directly on China's political system; his views are scattered in his comprehensive works on history, revolution, diplomacy and modernization, requiring extraction from his historical narrative and case studies.
\\\\
\textbf{DRAFT Answer:} John King Fairbank, a founding figure in Western Sinology, whose research on China's political system mainly includes two core perspectives: 1. Historical Continuity and Confucian Tradition: He emphasized understanding Chinese politics in a historical context, arguing that modern China's political system is deeply influenced by Confucian orthodoxy and imperial bureaucratic system..., with strong historical continuity in maintaining "great unity", centralization of power and social control mechanisms. 2. Modernization and the "Impact-Response" Model: In analyzing modern political system changes, he proposed the famous "impact-response" paradigm early on, holding that China's political transformation was mainly a passive response to Western Impact..., though this view was later revised and provided an important theoretical starting point for studying the adaptive adjustment of China's political system under external pressure.

\vspace{0.2cm}

\vpara{Example 5 (Information retrieval error):} 
\\\\
\textbf{Question:} Scholars engaged in gender studies in the Chinese mainland
\\\\
\textbf{Referenced Answer:} According to the knowledge base, the following are some scholars engaged in gender studies in the Chinese mainland and their research directions. Jin Xiaoyi is a professor at the School of Public Policy and Administration, Xi’an Jiaotong University, and her research directions include gender imbalance, marriage squeeze, gender norms, etc. Yang Xueyan is a professor at the School of Public Policy and Administration, Xi’an Jiaotong University, with research directions covering gender imbalance and social stability, gender innovation, etc. Tong Xin is a professor at the Department of Sociology, Peking University, and her research directions include labor sociology, gender studies, women’s career development, etc. Hu Yukun is a researcher at the Institute for Global Health and Development, Peking University, focusing on social gender and health, women’s reproductive health, etc. Zhang Meifang is a professor at the Institute of History of Science and Cultural Heritage, University of Science and Technology Beijing, and her research directions involve gender and science studies, science and technology culture communication, etc. These scholars come from multiple disciplines including sociology, public administration, history of science and technology, and public health, which reflects the interdisciplinary nature of gender studies. Their research topics cover important fields in Chinese society such as gender imbalance, gender norms, women’s career development and health.
\\\\
\textbf{DeepSeek Answer:} According to the provided knowledge base information, for the query of "scholars engaged in gender studies in the Chinese mainland", the system returns an empty author list and no eligible scholar records are found. This indicates that no relevant information can be retrieved under the current query parameters. Potential causes include the following aspects: on the one hand, in terms of data coverage, the knowledge base may not have collected or indexed the data of scholars whose research interest is "gender studies" and nationality is "Chinese"; on the other hand, restricted by query parameters, the query may be set with specific sorting (such as descending order by citation count) or result quantity limits, leading to no matching items in the existing data set. It is suggested that to obtain relevant information, attempts can be made to adjust the search conditions, for example, relaxing the matching scope of the "interest" field, or using other academic databases for supplementary retrieval.

\section{Academic Q\&A Benchmark Annotation Check System}

This system is a dedicated annotation and validation platform designed to ensure the quality of the \textit{Academic Q\&A Benchmark} dataset, supporting the development and evaluation of academic question-answering models. It enhances the reliability and utility of benchmark data by enabling human annotators to systematically review, verify, and refine automatically generated answers.

\subsection{Core Functionality}
\begin{itemize}[leftmargin=*]
    \item \textbf{Question Navigation \& Multi-dimensional Annotation}: Annotators can quickly locate questions by serial number and classify them along key dimensions, including question retention, difficulty, API solvability, search entity type, and cognitive level.
    \item \textbf{Answer Review \& Validation}: The system aggregates automatically generated content, including API planning steps, input/output data, and natural language answers. Annotators validate answer correctness, edit incorrect responses, and provide review comments and reference links.
    \item \textbf{Progress Management}: All inspection records and timestamps are tracked, with navigation controls to streamline the review workflow.
\end{itemize}

\subsection{Value}
By combining automated answer generation with \textit{human-in-the-loop validation}, the system ensures the dataset contains accurate, relevant, and well-supported answers, providing a robust foundation for training and evaluating next-generation academic question-answering systems.

\section{\benchname Annotation  System}
\label{sec:anno_system}

This system is a dedicated platform for \textbf{API-driven annotation and debugging} within the Academic Q\&A Benchmark, designed to support the construction of high-quality, explainable academic question-answering (QA) data. It enables annotators to design, execute, and validate API-based solution workflows for complex scholarly queries.

\subsection{Core Functionality}
\begin{itemize}[leftmargin=*]
    \item \textbf{API Call Planning}: Annotators can visually design sequences of API calls (e.g., \texttt{search\_author\_id}, \texttt{search\_author\_detail}), defining dependencies, execution order, and parameters (e.g., scholar name, institution) to solve the target question.
    \item \textbf{One-Click Execution \& Debugging}: The system supports step-by-step or one-click execution of the planned API workflow, allowing annotators to inspect parameters, view raw return results, and refine the plan.
    \item \textbf{Result Evaluation}: After execution, annotators evaluate the generated answer across multiple dimensions: overall quality, coverage of relevant information, and accuracy of retrieved data, while also providing justifications and reference links for their evaluations.
    \item \textbf{Progress Tracking}: The system records all annotation activities (including timestamps and historical annotations) and supports callback to specific issues for iterative refinement.
\end{itemize}

\subsection{Value}
By integrating API planning, execution, and human-in-the-loop evaluation, this system ensures that the benchmark dataset contains not only accurate answers but also transparent, reproducible reasoning paths, providing a robust foundation for training and evaluating next-generation academic QA models that can effectively leverage external knowledge sources.

\hide{
\section{Academic Q\&A Benchmark API Debugging \& Validation Platform}

This platform is an integrated tool for \textbf{API-driven debugging and answer validation} within the Academic Q\&A Benchmark, supporting the construction of high-quality, reproducible academic question-answering (QA) data.

\subsection{Core Functionality}
\begin{itemize}[leftmargin=*]
    \item \textbf{API Debugging}: Provides documentation for academic knowledge APIs (e.g., \texttt{search\_paper\_id}, \texttt{search\_author\_id}), including function descriptions, request parameters, and response examples. It allows editing JSON parameters, validating syntax, and executing API calls to debug solution workflows.
    \item \textbf{Answer Validation}: Integrated with the annotation system, it enables human reviewers to verify answer correctness, provide comments, and attach reference links, ensuring benchmark data reliability.
\end{itemize}

\subsection{Value}
By combining **API debugging and human-in-the-loop validation**, the platform ensures the benchmark dataset contains accurate answers and transparent reasoning paths, offering a robust foundation for training and evaluating next-generation academic QA models.

\begin{figure*}[t]
\centering
\includegraphics[width=0.98\linewidth]{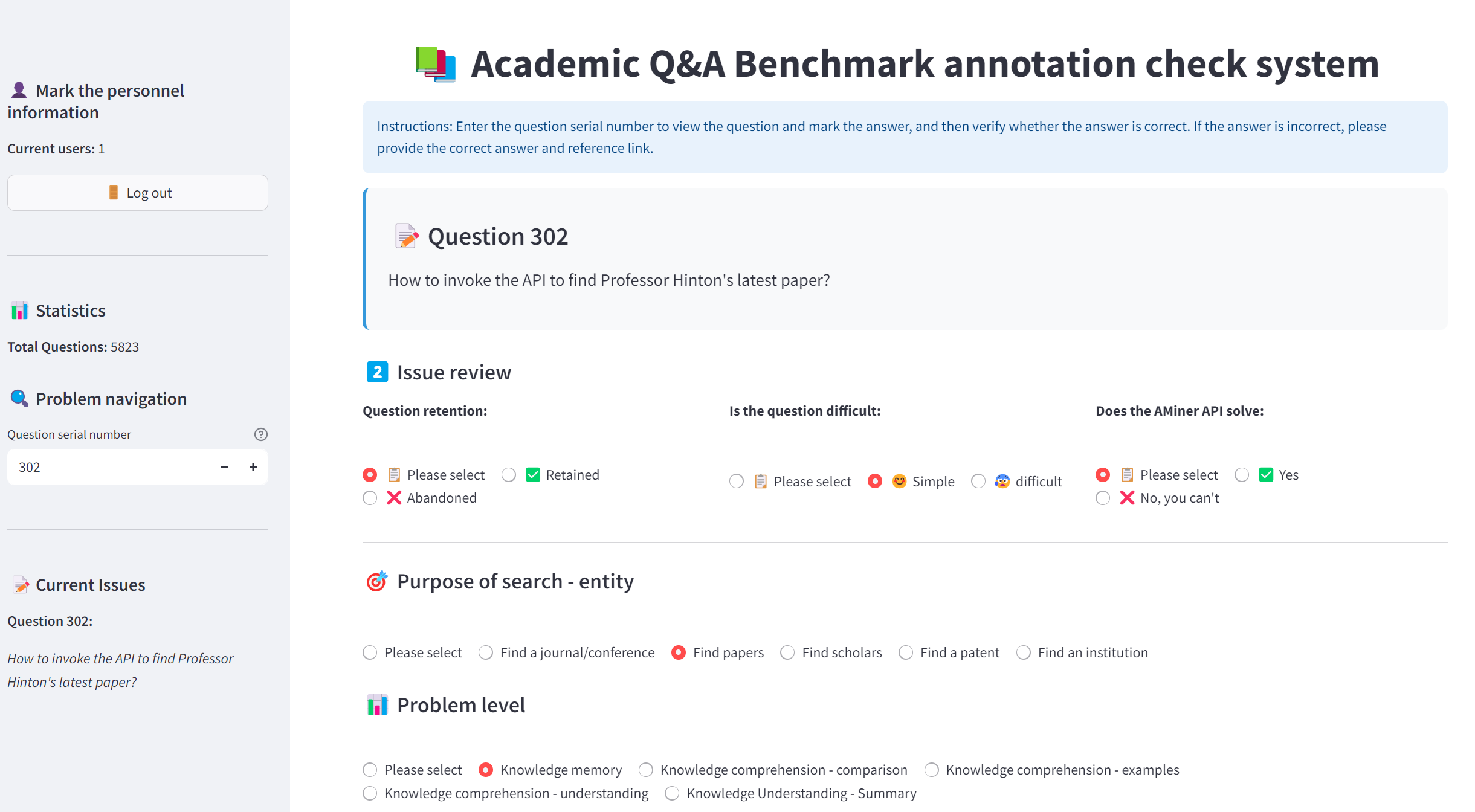} 
\caption{Screenshot of the Review Interface 1}
\vspace{1cm}
\label{fig:figure 8}
\end{figure*}

\begin{figure*}[t]
\centering
\includegraphics[width=0.98\linewidth]{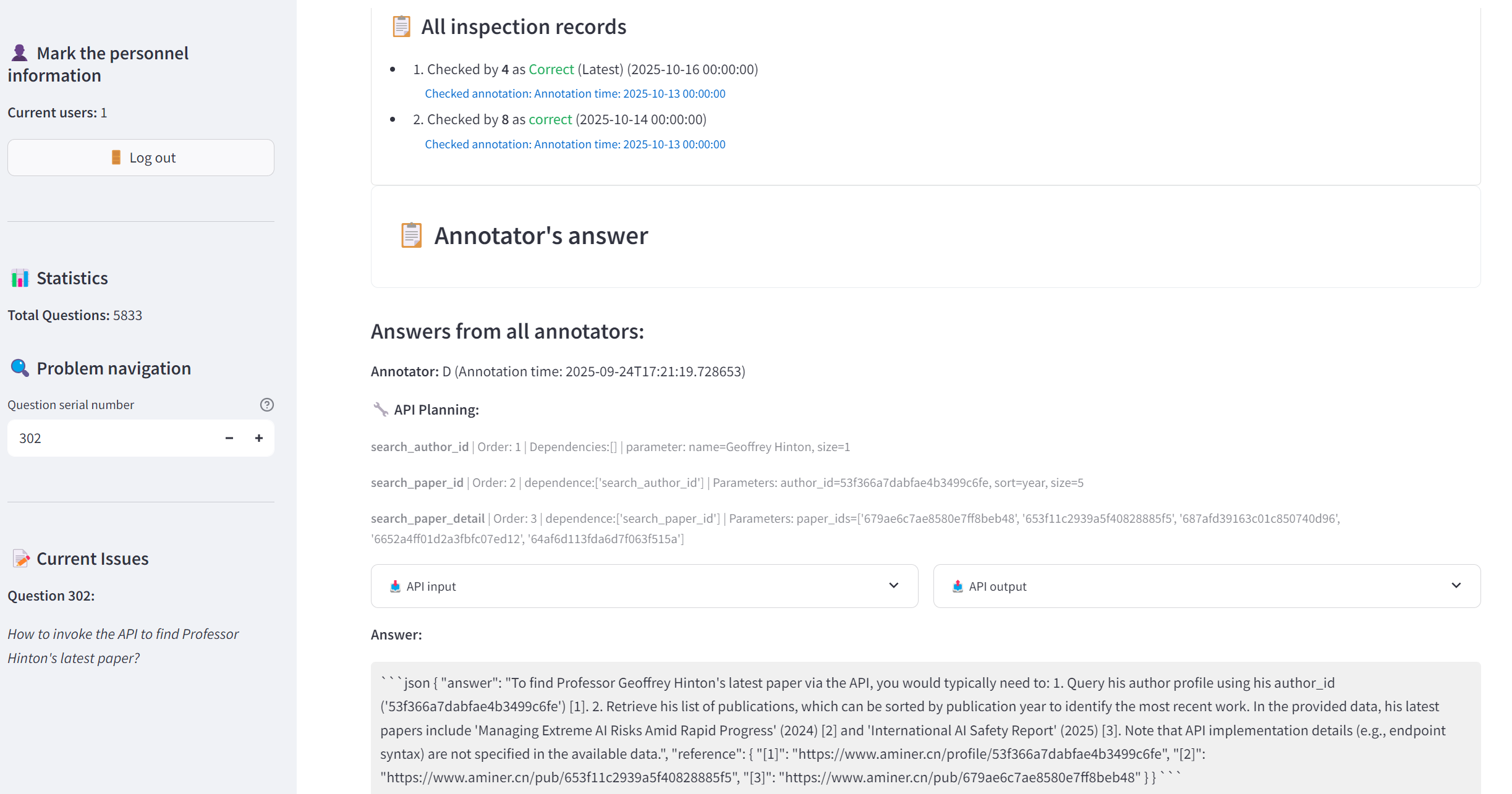} 
\caption{Screenshot of the Review Interface 2}
\label{fig:figure 9}
\end{figure*}

\begin{figure*}[t]
\centering
\includegraphics[width=0.98\linewidth]{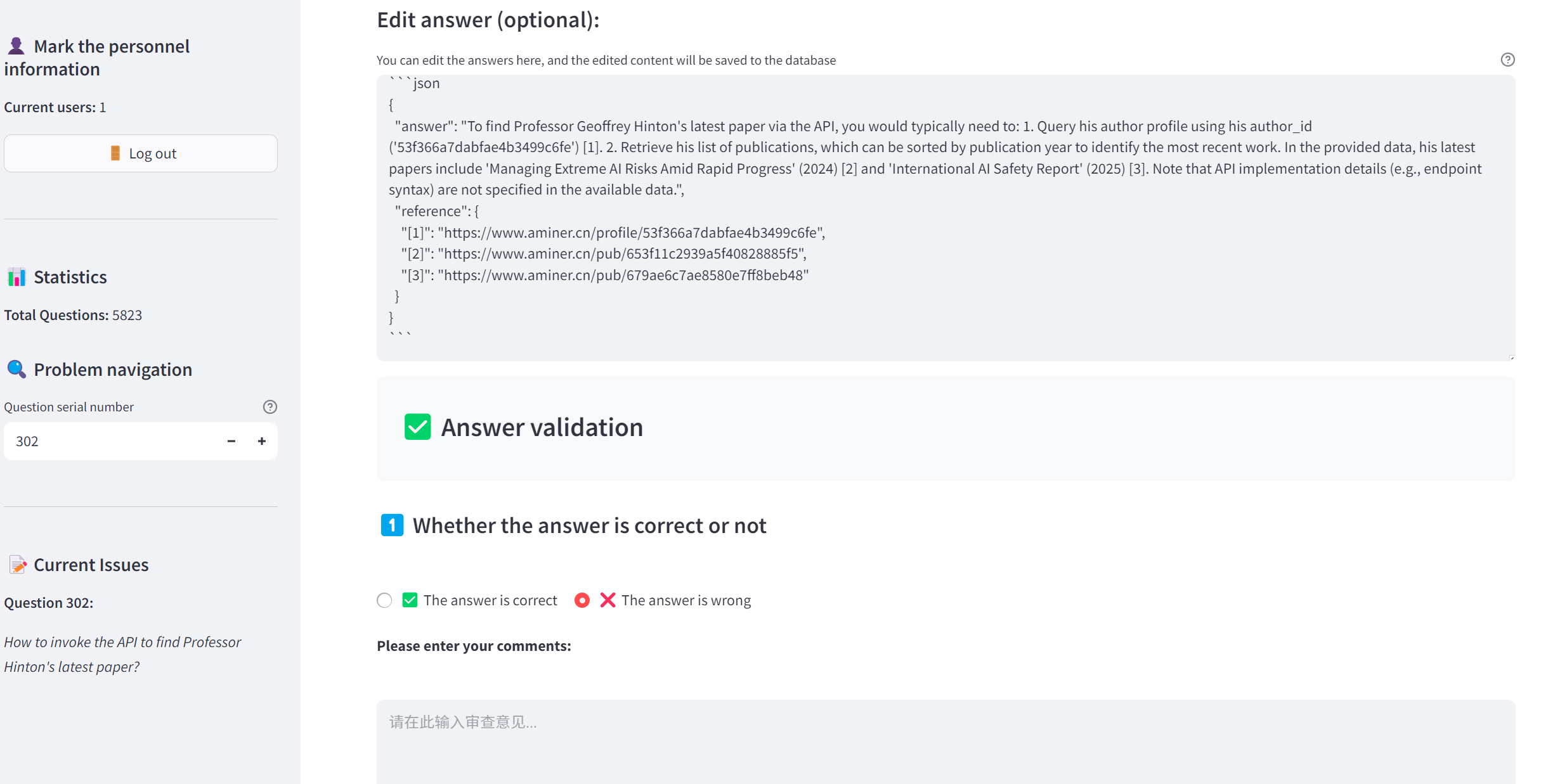} 
\caption{Screenshot of the Review Interface 3}
\vspace{1cm}
\label{fig:figure 10}
\end{figure*}

\hide{
\begin{figure*}[t]
\centering
\includegraphics[width=0.98\linewidth]{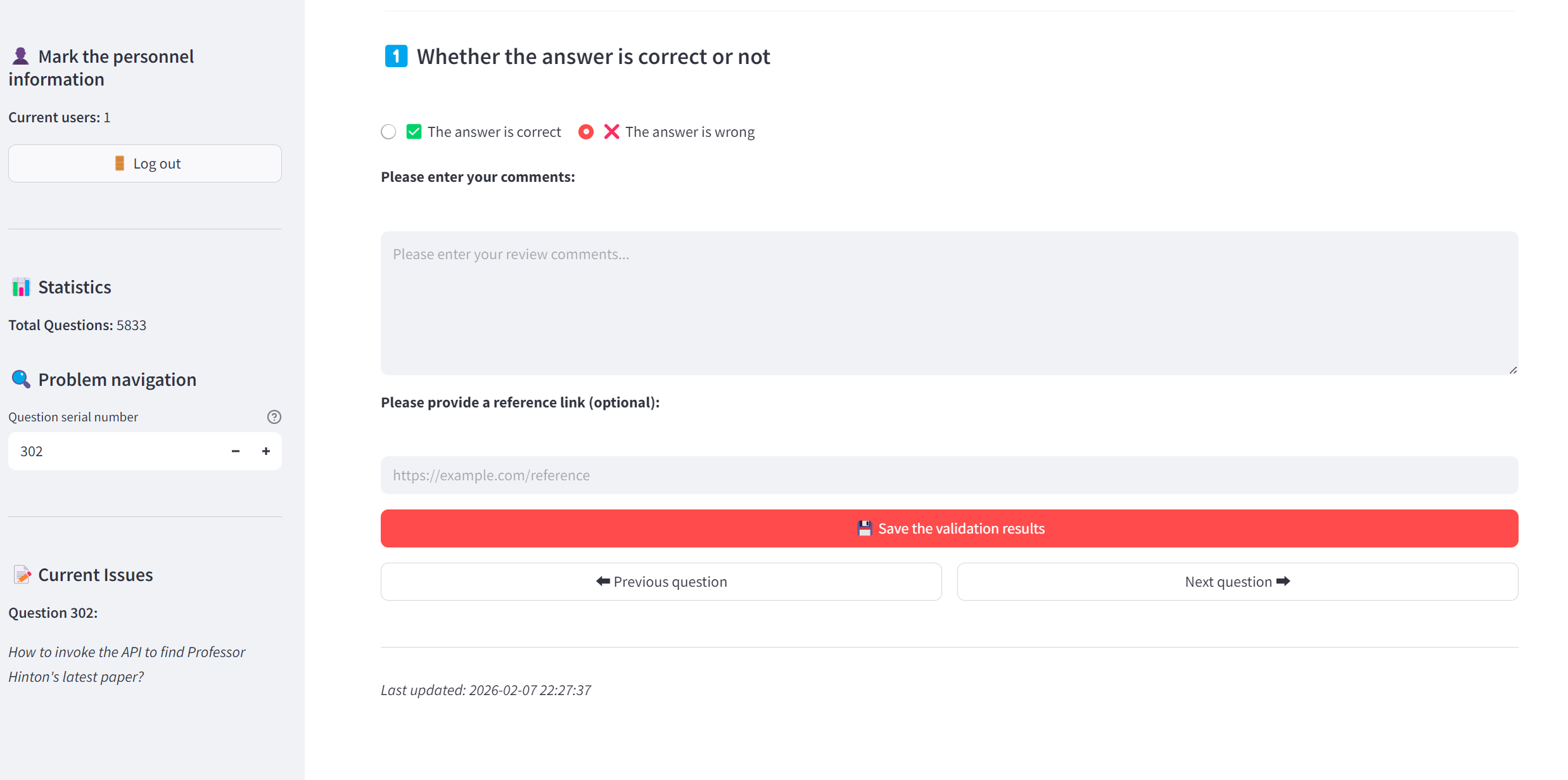} 
\caption{Screenshot of the Annotation Interface 4}
\label{fig:figure 11}
\end{figure*}
}

\begin{figure*}[t]
\centering
\includegraphics[width=0.98\linewidth]{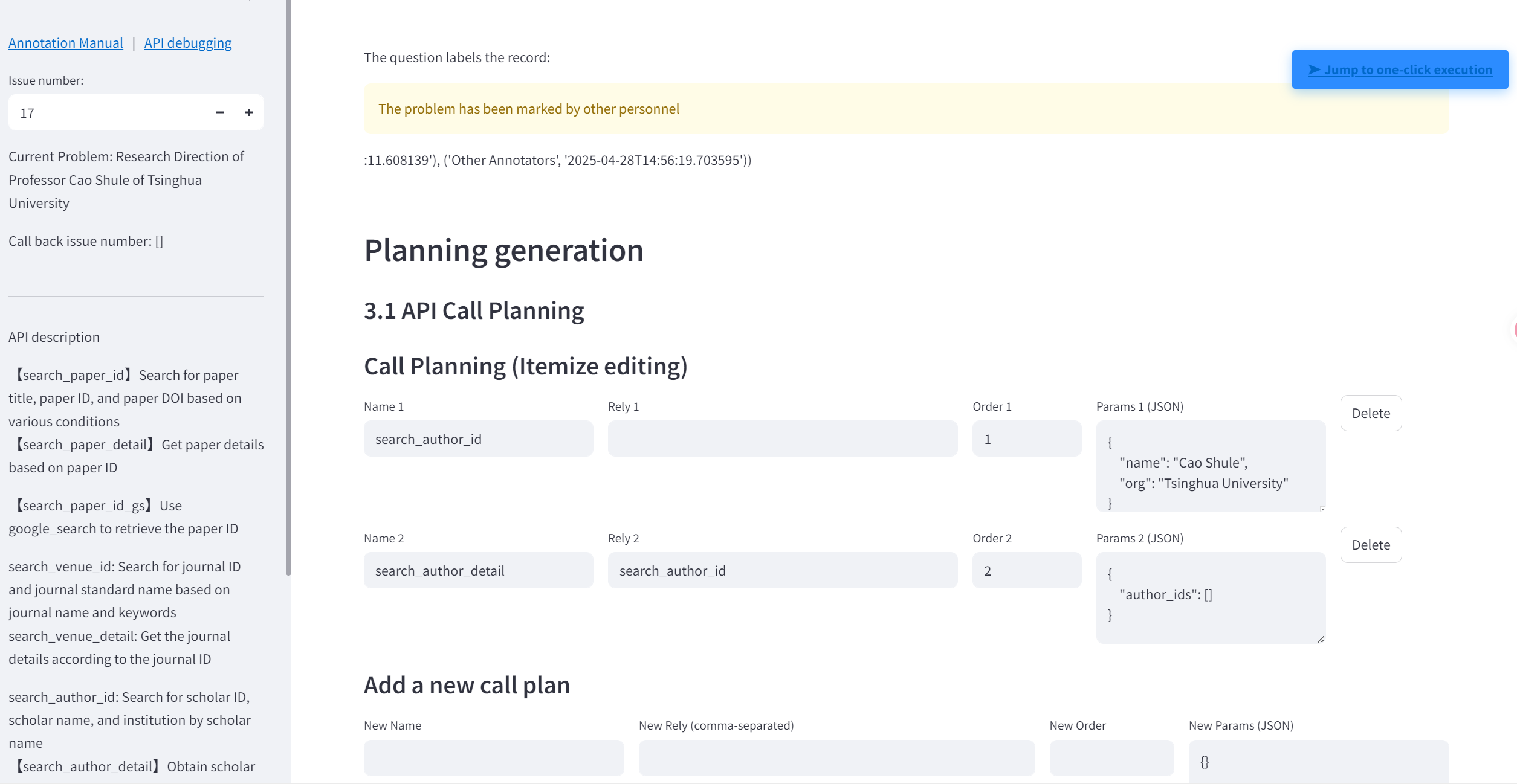} 
\caption{Screenshot of the Annotation Interface 1}
\vspace{1cm}
\label{fig:figure 12}
\end{figure*}

\begin{figure*}[t]
\centering
\includegraphics[width=0.98\linewidth]{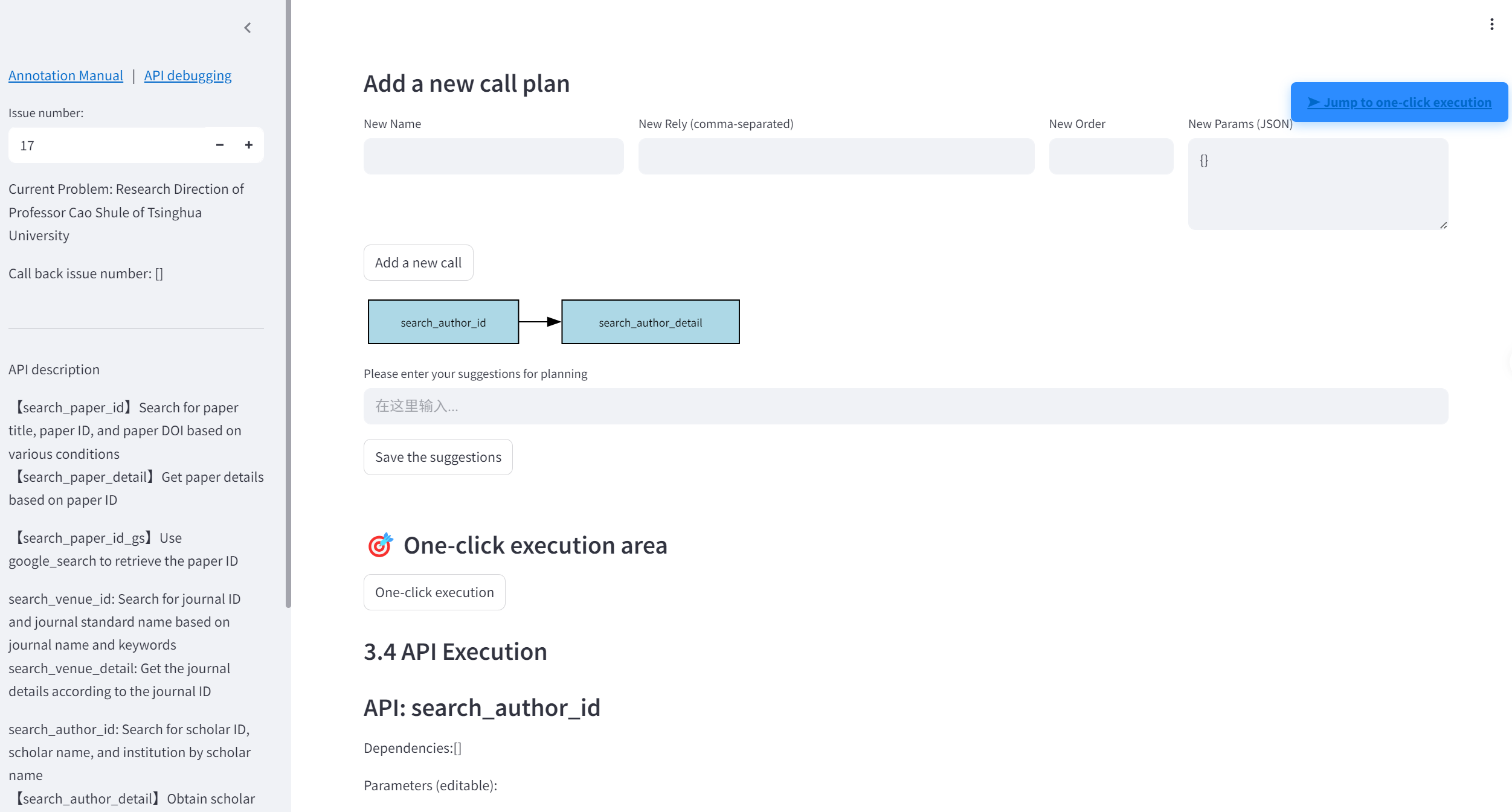} 
\caption{Screenshot of the Annotation Interface 2}
\vspace{1cm}
\label{fig:figure 13}
\end{figure*}

\begin{figure*}[t]
\centering
\includegraphics[width=0.98\linewidth]{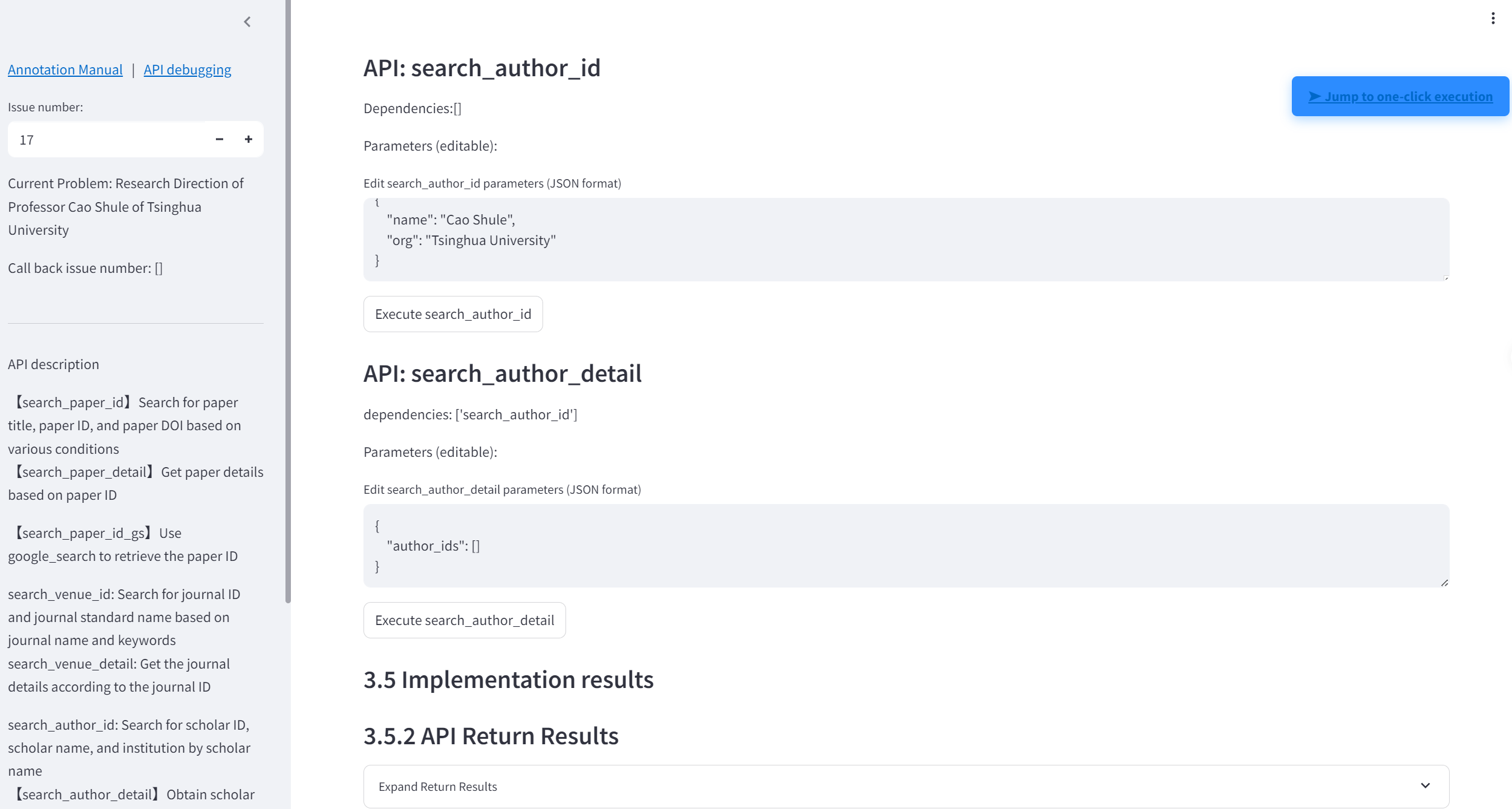} 
\caption{Screenshot of the Annotation Interface 3}
\vspace{0.2cm}
\label{fig:figure 14}
\end{figure*}

\begin{figure*}[t]
\centering
\includegraphics[width=0.98\linewidth]{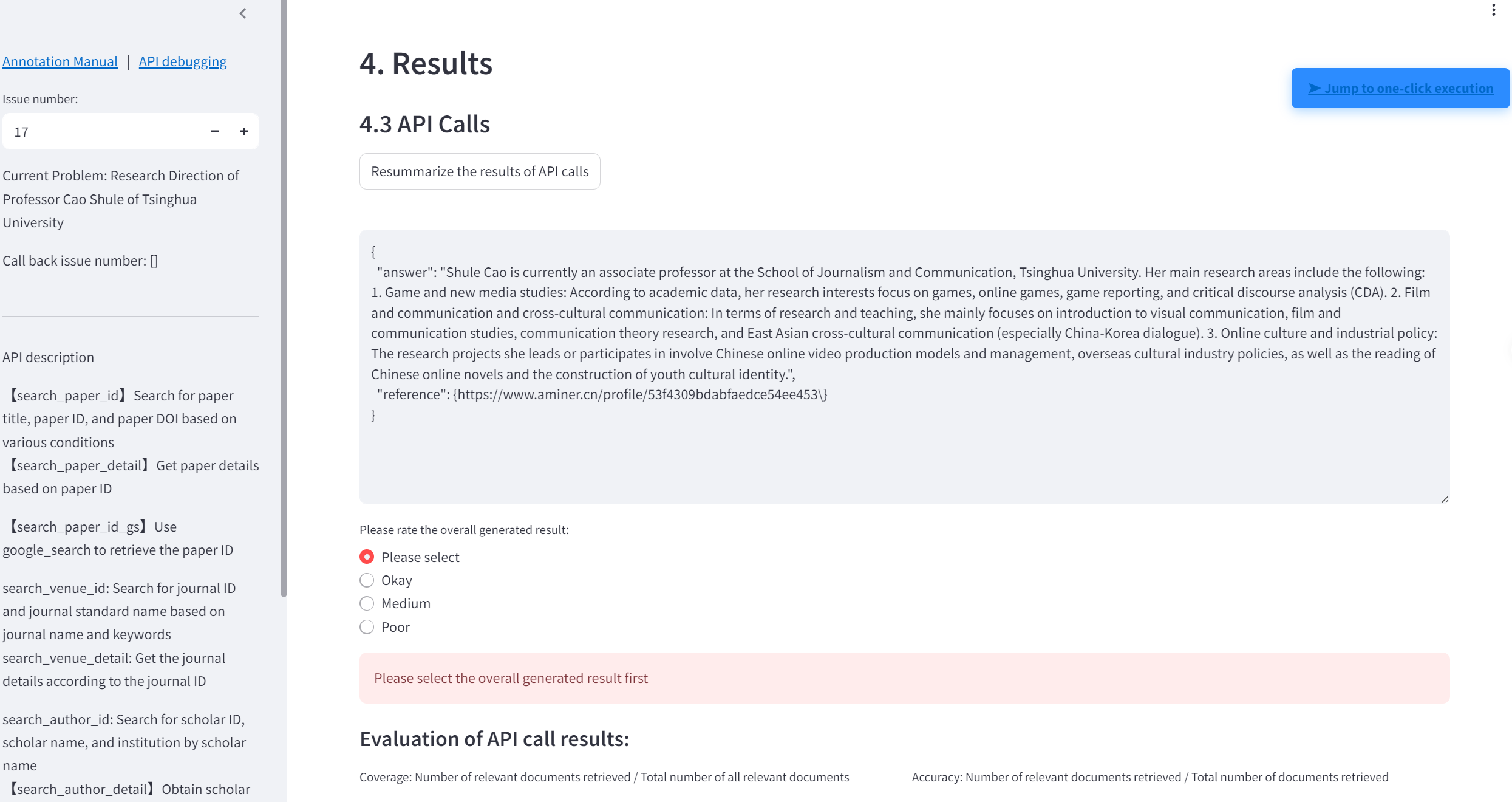} 
\caption{Screenshot of the Annotation Interface 4}
\vspace{1cm}
\label{fig:figure 915}
\end{figure*}

\hide{
\begin{figure*}[t]
\centering
\includegraphics[width=0.98\linewidth]{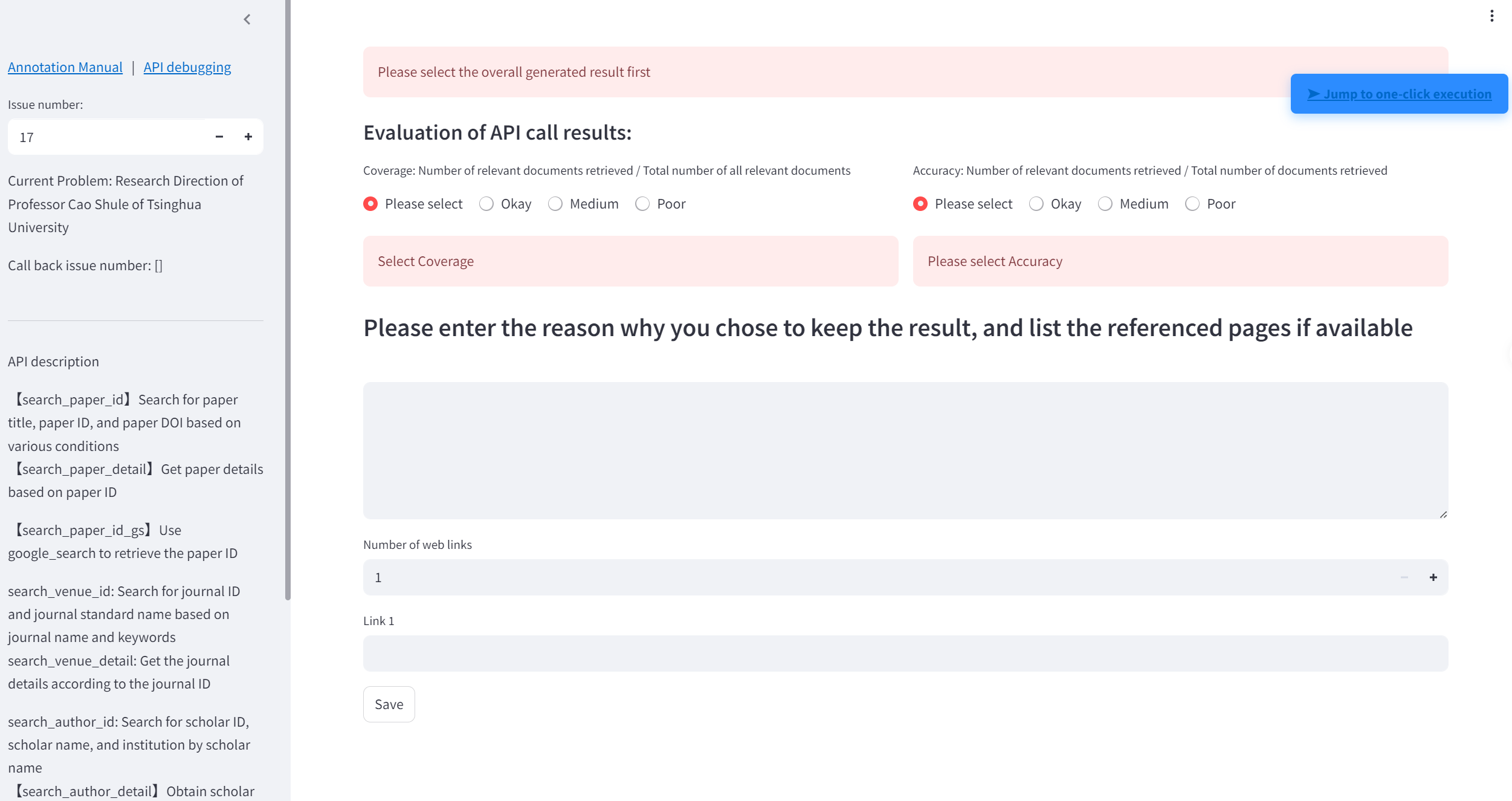} 
\caption{Screenshot of the Review Interface 5}
\label{fig:figure 16}
\end{figure*}
}

\begin{figure*}[t]
\centering
\includegraphics[width=0.98\linewidth]{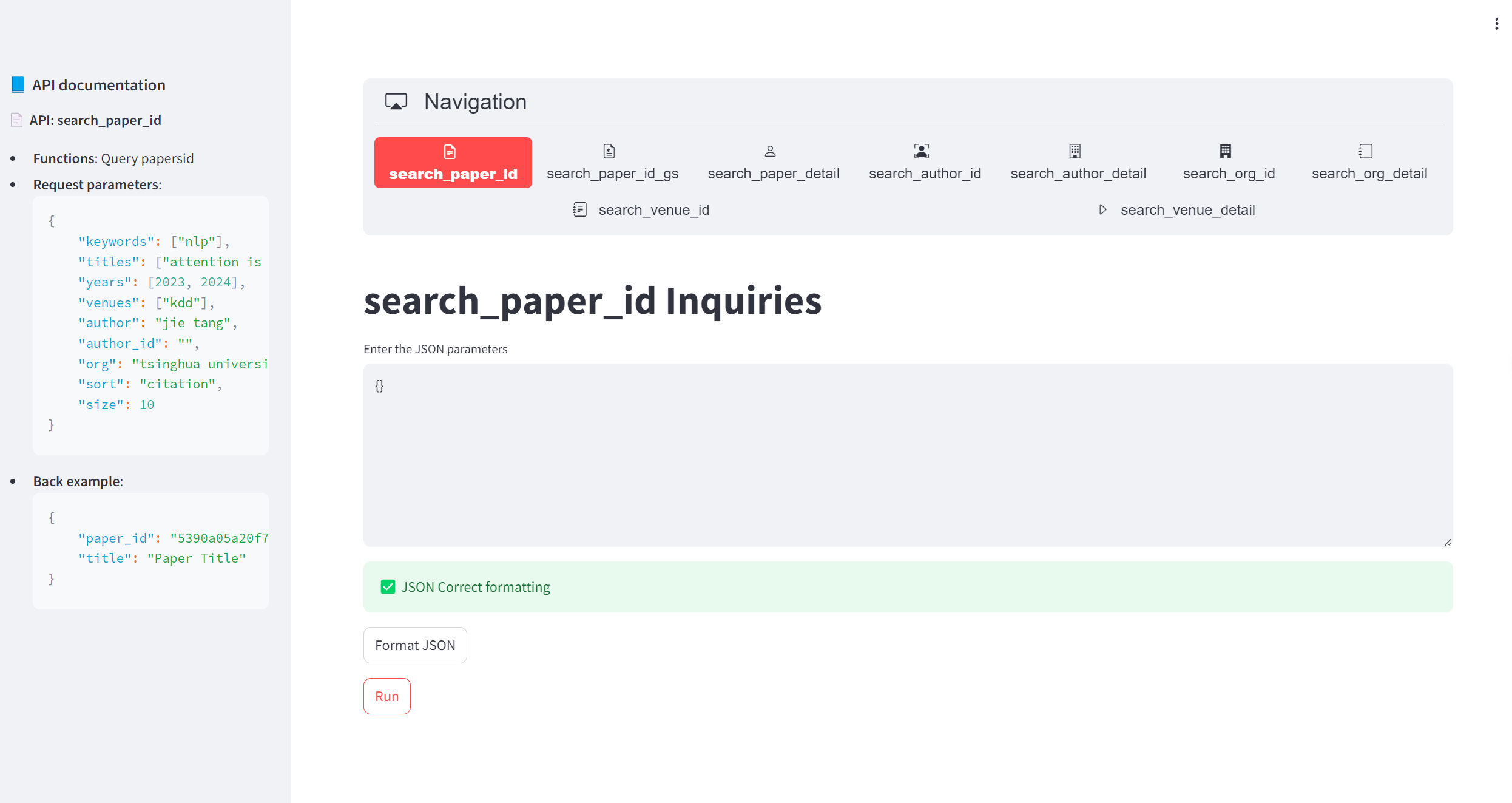} 
\caption{Screenshot of the API Debugging Interface}
\vspace{0.5cm}
\label{fig:figure 17}
\end{figure*}

\hide{
\begin{figure*}[t]
\centering
\includegraphics[width=0.98\linewidth]{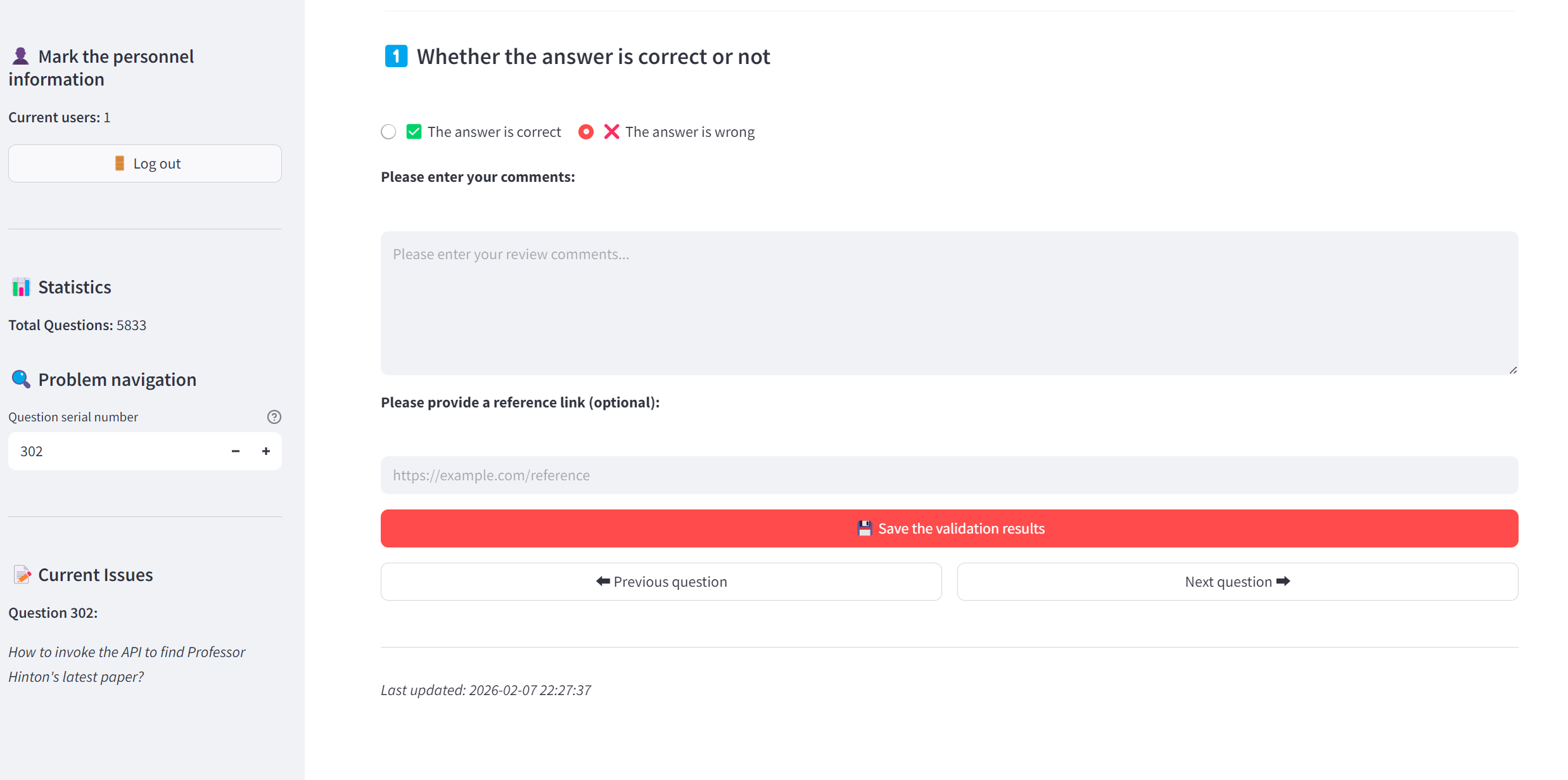} 
\caption{Screenshot of the API Debugging Interface 2}
\label{fig:figure 18}
\end{figure*}
}

\section{Decompose Prompt}

\subsection{Prompts for CAW}
~\\
\vpara{Result Prompt}
\begin{lstlisting}[style=prompt]
"""
You are a world-class academic expert. You will answer questions from any academic field.

Instructions:

Your answer must be in the same language as the question. 

The answer must be concise and scholarly, no longer than 300 words.

Analyze the user's question to determine the underlying intent(s). For each distinct intent, provide a separate, clearly labeled point in your answer.

Use the following knowledge base information (api_output) as your primary source. Only use data from the knowledge base that is directly relevant to the question and its intent(s). Do not use unrelated information, even if present in the knowledge base.

If your answer contains any information drawn from api_output (even partially), you must include a citation [e.g., [1], [2]] for that content. If the information is insufficient, you may supplement with your own knowledge, but always prioritize and cite api_output when used.
In the "reference" field, each citation label (e.g., [1], [2]) should map to a full citation (with the appropriate link, as specified above).
In the "reference" field, only use links or IDs that are present in the knowledge base (api_output); do not invent, infer, or fabricate any links or IDs.
In the "reference" field, citations must be numbered sequentially as [1], [2], [3], [4], etc., starting from [1] and increasing by 1 for each new citation. No duplicate citations are allowed in the "reference" field; each cited item should appear only once in the reference list.

If a citation is an ID, determine its type (paper, author, venue, or org) based on the knowledge base (api_output), and generate the corresponding full URL as follows:
- For papers: https://www.aminer.cn/pub/paper_id
- For scholars: https://www.aminer.cn/profile/author_id
- For journals/conferences: https://www.aminer.cn/open/journal/detail/venue_id
- For institutions: https://www.aminer.cn/institution/org_id
Only generate such URLs if the ID exists in the knowledge base (api_output); do not fabricate or infer IDs.

Output only a JSON object with two fields:
- "answer": your academic answer, written in the same language as the question (max 300 words).
- "reference": a dictionary mapping each citation label to its bibliographic citation (as a webpage or ID from the knowledge base).

Example output:
{{
"answer": "Your detailed answer here, with inline citations [1][2] ...",
"reference": {{"[1]": "https://xxx", "[2]": "ID"}}
}}

Here is the knowledge base output (api_output):
{api_output}

Here is the question:
{question}
"""
\end{lstlisting}

\vpara{Plan Prompt}
\begin{lstlisting}[style=prompt]
"""
You are a planning expert. Based on the user's question, you will select one or more appropriate APIs and retrieve relevant information through them in order to accurately answer the user's question. [The thinking process should not exceed ten sentences.]
   The output format you generate should follow this example:
    [
        {
            "name": "search_author_id",
            "rely": [],
            "order": 1,
            "params": {"interest": ["Optical communication"]}
        },
        {
            "name": "search_author_detail",
            "rely": ["search_author_id"],
            "order": 2,
            "params": {"ids": []}
        },
    ]
    # Description: name refers to the API name; rely indicates the source API names for input parameters (must be API names only); order indicates execution order such as 1, 2, 3; params refers to API parameters.
    # Note: You do not need to generate any extra explanations-only output the JSON content described above! (Do not include words like ``` or json.)
    # Note: If parameters do not have specific values, use "" empty strings, empty lists, or other empty values.
    # Note: true and false must be lowercase!
    # Note: If no sorting method is specified, always sort by citation!
    # Note: If multiple APIs share the same name, add numbering in the name field such as search_paper_id(1), search_paper_id(2), etc.
    # Note: If the question is not specified as Chinese, all keywords must be English words!

    # Note: You are not limited to the planning order above; you may design any API calling order based on your knowledge!

    # Available APIs are as follows:
[search_paper_id]
    "search_paper_id": {
        "description": "Search paper IDs based on conditions",
        "parameters": {
            "titles": ["paper titles"], 
            "keywords": ["keywords"], 
            "years": {
                "type": "array",
                "description": "List of publication years, represented as integers"
            },
            "is_sci": {
                "type": "boolean",
                "description": "Whether it is an SCI paper"
            },
            "language": "Paper language (str), using ISO 639-1 codes such as 'en', 'zh'",
            "sort": "Sorting method (str), must be one of: year, citation",
            "author": "Author name (str)",
            "author_id": "Author ID (str)",
            "coauthors": {
                "type": "array",
                "description": "List of co-author names, excluding the author specified in the author parameter"
            },
            "org": "Institution or university name (str)",
            "org_id": "Institution or university ID (str)",
            "venues": {
                "type": "array",
                "description": "List of journals or conferences, using lowercase English abbreviations without years"            
            },
            "venue_ids": ["journal or conference IDs"], 
            "size": {
                "type": "integer",
                "description": "Number of results returned, must be less than 100"                   
            }
        },
        "response": {
            "description": "List of results, each element is a dictionary representing a paper",
            "data": [
                {
                    "paper_id": "paper ID (str)",
                    "title": "paper title (str)"
                }
            ]
        }                     
    },

---
[search_paper_detail]
    "search_paper_detail":{
        "description": "Retrieve detailed paper information based on paper ID list",
        "parameters": {
            "paper_ids": ["paper IDs"]
        },
        "response":{
            "description": "List of results, each element is a dictionary representing a paper",
            "data": [
                {
                    "paper_id": "paper ID (str)",
                    "title": "paper title (str)",
                    "abstract": "paper abstract (str)",
                    "year": "publication year (float)",
                    "citation": "citation count (float)",
                    "keywords": ["keywords"],
                    "authors": [
                        {
                            "author": "author name (str)",
                            "author_id": "author ID (str)",
                            "org": "author institution (str)",
                            "org_id": "author institution ID (str)",
                            "email":"author email (str)"
                        }
                    ],
                    "org": "publishing institution (str)",
                    "org_id": "publishing institution ID (str)",    
                    "venue": "journal or conference (str)"
                }
            ]
        }
    },

---
[search_author_id]
    {
        "type": "function",
        "function":{
            "name": "search_author_id",
            "description": "Search scholars based on name, institution, interests, country, etc.",
            "parameters": {
                "type": "object",
                "properties": {
                    "name": {
                        "type": "string",
                        "description": "Scholar name"
                    },
                    "org": {
                        "type": "string",
                        "description": "Scholar institution"
                    },
                    "size": {
                        "type": "integer",
                        "description": "Number of scholars returned, maximum 1000"
                    },
                    "interest": {
                        "type": "list",
                        "description": "Scholar interests, format [str,str,...]"
                    },
                    "nation": {
                        "type": "list",
                        "description": "Scholar countries, format [str,str,...]"
                    },
                    "order": {
                        "type": "string",
                        "description": "Sorting field: n_citation, n_pubs, h_index"
                    },
                    "asc": {
                        "type": "boolean",
                        "description": "true ascending, false descending"
                    }
                },
                "required": []
            }
        }
    },
---
[search_author_detail]Retrieve scholar details based on scholar ID list
Input: ids:[str,str, ...]  # ID list

---
[search_venue_id]Search journal ID and standardized name based on journal name
Input: 
    name:str  # journal name
    category:str  # discipline category name
    category_source:string # numeric string indicating source: 0:"SJR",1:"WOS",2:"GB09",3:"CCF",4:"CSCD",5:"CCJ",6:"ARXIV",7:"CJCR",8:"JCR",9:"SCI"
    quartile:str  # quartile search such as "Q1","A"
    keywords:list  # journal keyword list
    size:number  # number of journals returned
Output:journal id

---
[search_venue_detail]Retrieve journal details based on journal IDs
Input:
    ids:[str,str, ...]  # ID list
Output:
alias	array	aliases
category_id	string	category id
classes	array	data sources
id	string	id
issn	string	ISSN
lower_alias	array	aliases lowercase
name	string	name
name_en	string	English name
name_zh	string	Chinese name
num	float	priority number
quartile	string	quartile
source_quartiles	array	source quartiles
total	float	total count
type	string	category system
url	string	source url    

---
[search_org_id]Search institution IDs based on name keywords
Input:
    orgs:[str,str,...]  # institution names
Output:
    institution id list

---
[search_org_detail]Retrieve institution details based on institution IDs
Input:  
    ids:[str,str, ...]  # ID list
Output:
acronyms	array	
aliases	array	
coordinate	array	
details	array	
error	array	
established	int	
external_ids	array	
geographic_id	string	
id	string	
image	string	
introduction	string	
language	string	
latitude	float	
longitude	float	
name	string	
name_en	string	
name_zh	string	
relationships	array	
src	string	
total	int	
type	string	

[search_paper_id_gs]Retrieve paper IDs using Google Scholar
Input:  
    query: "str" # user question
"""
\end{lstlisting}

\vpara{Dynamic Parameter Prompt}
\begin{lstlisting}[style=prompt]
"""
Please regenerate the parameters required for {api_name} based on the data provided by the user.
The parameter format example is as follows: {PARA}
The available parameters for the API are as follows: {params_temp}
The reference parameters are as follows: {params}
You need to identify the parameters required for {api_name} from these, and generate them according to the format example.
[Please note, you only need to generate the parameter format above, no additional explanations or comments are needed!]
[You need to generate the complete parameters, do not truncate in the middle!]
"""
\end{lstlisting}

\subsection{Prompts for Evaluation Metrics}
~\\
\vpara{Correctness Prompt}
\begin{lstlisting}[style=prompt]
"""
Please score the "Correctness" of the answer to be evaluated, with a rating range of [0,1]:

Question: {question}
Reference Answer: {gold}
Answer to be Evaluated: {pred}

Definition of Correctness: Compared with the reference answer, whether the evaluated answer accurately addresses the question. You may estimate it as (relevant information appearing in the evaluated answer / all information appearing in the evaluated answer).
Very Poor - All content in the evaluated answer is irrelevant to the reference answer and the question.
Poor - The evaluated answer contains a large amount of irrelevant content.
Fair - The evaluated answer contains some irrelevant information.
Good - The evaluated answer is mostly relevant, with only minor deviations.
Excellent - All content in the evaluated answer is relevant to the reference answer and the question.

Note: If the evaluated answer only provides search steps or methods but does not provide actual results, the score must be 0.

**Mandatory Scoring Rules:**
1. **List/Person/Paper-type Questions** (e.g., "Who are the collaborators", "Which researchers", "Related papers"):
   First compute the **matching percentage** = number of items in the evaluated answer that are consistent with the reference answer (same person, same paper, or same link counts as consistent) / total number of items in the reference answer.
   The rating should follow this percentage (e.g., 4 correct out of 5 gives 4/5 = 0.8; 100% -> 1.0, 0% -> 0).
   If the evaluated answer has **no overlap at all** with the reference answer, the score must not exceed **0.5**.
2. **Paper/Literature Questions** (e.g., "related papers", "which publications"):
   If the evaluated answer **does not provide any paper links** (reference is empty, or claims retrieval failure, or no data returned), the accuracy must not exceed **0.8**.
   If links are provided, scoring should still follow the matching-percentage rule above.
3. For non-list questions (e.g., factual judgment or single conclusion), score based on overall relevance and correctness compared to the reference answer.

Output JSON:
{{"rating": number}}
"""
\end{lstlisting}

\vpara{Integrality Prompt}
\begin{lstlisting}[style=prompt]
"""
Please score the "Integrality" of the answer to be evaluated, with a rating range [0,1]:
1 means fully covered, 0 means not covered at all.

Question: {question}
Reference Answer: {gold}
Answer to be Evaluated: {pred}

Definition of Integrality: Compared with the reference answer, whether the evaluated answer is complete in terms of including all expected components and sub-items. You may estimate it as (number of covered components / total components in reference answer).
Very Poor - All components are missing.
Poor - Most components are missing.
Fair - Some components are missing.
Good - Most components are covered, with only minor deviations.
Excellent - All components in the reference answer are fully included.

Note: If the evaluated answer only provides search steps or methods but does not provide actual results, the score must be 0.

Output JSON:
{{"rating": number}}
"""
\end{lstlisting}

\vpara{Completeness Prompt}
\begin{lstlisting}[style=prompt]
"""
Please evaluate whether the answer has "fulfilled the core objective of the question and is factually correct", with a rating range of [0,1]:
1 = fully covered and correct, 0 = not fulfilled or incorrect.
**Completion considers both coverage and correctness: whether all sub-queries/checkpoints are addressed, and whether each response is consistent with and correct relative to the reference answer.**

Question: {question}
Reference Answer: {gold}
Answer to be Evaluated: {pred}

You should evaluate as follows:
(1) Extract all **sub-queries** from the user query.
(2) Determine the **central intent** for each sub-query.
(3) Based on the reference answer, extract an **evaluation checklist** corresponding to each sub-query's central intent.

Please strictly follow these guidelines:
1. **Sub-query Identification**
   - If the user query has only one core intent, extract the full query as the single sub-query.
   - If the user query contains multiple core intents, extract each as an independent sub-query.
   - DO NOT fabricate non-existent sub-queries or extract irrelevant minor details.
2. **Sub-query Rewriting**
   - If there is only one core intent, keep the original query unchanged.
   - If there are multiple intents, rewrite each sub-query into an independently understandable question.
3. **Central Intent**
   - Identify the expected final output form for each sub-query (e.g., paper list, researcher names, factual conclusions, research viewpoints).
   - Ignore intermediate reasoning steps and focus only on the user's final desired outcome.
4. **Evaluation Checklist Extraction**
   - Extract key factual evaluation points from the reference answer that directly correspond to the central intent.
   - Each factual entity (person, paper, institution, attribute, conclusion, etc.) must be listed as an independent item.
   - The checklist should be sufficient for evaluation without needing to read the full reference answer.
   - Ignore reference URLs; only consider factual textual content.

During reasoning, generate sub-queries, rewritten sub-queries, central intents, and evaluation checklists, and then assign a score based on how completely and correctly the evaluated answer satisfies the checklist (coverage without correctness must be penalized).

Output:
{{"rating": number}}
"""

\end{lstlisting}

\vpara{Faithfulness Prompt}
\begin{lstlisting}[style=prompt]
"""
Please score the "Faithfulness" of the answer to be evaluated, i.e., whether it strictly follows the API outputs without fabrication. Rating range [0,1]:
Faithfulness definition: Whether the answer is strictly based on API-returned content and contains no fabricated or hallucinated information.
1 = completely faithful, 0 = completely fabricated.

Answer: {pred}
API Output: {api_output}

Output JSON:
{{"rating": number}}
"""
\end{lstlisting}
}

\end{document}